\documentclass[11pt,a4paper]{article}

\usepackage[utf8]{inputenc}
\usepackage[T1]{fontenc}
\usepackage{geometry}
\usepackage{graphicx}
\usepackage[table]{xcolor}
\usepackage{fontawesome5}
\usepackage{hyperref}
\usepackage{titlesec}
\usepackage{fancyhdr}
\usepackage{amsmath}
\usepackage{amsfonts}
\usepackage{amssymb}
\usepackage{booktabs}  
\usepackage{array}  
\usepackage{algorithm}
\usepackage{algorithmic}
\usepackage{tabularx}
\usepackage{fvextra}
\usepackage{appendix}

\usepackage[most]{tcolorbox}
\usepackage{listings}
\usepackage{subcaption}
\usepackage{float}

\usepackage{tocloft}

\lstnewenvironment{codebox}[2][]
{
  \lstset{
    basicstyle=\ttfamily\small,
    columns=fullflexible,
    keepspaces=true,
    breaklines=true,
    showstringspaces=false,
    numbers=left,
    numberstyle=\tiny\color{gray},
    stepnumber=1,
    numbersep=8pt,
    backgroundcolor=\color{black!2},
    frame=single,
    frameround=tttt,
    rulecolor=\color{black!45},
    captionpos=t,
    title={#1},
    language=#2
  }
}
{}

\definecolor{linkblue}{RGB}{0,102,204}

\hypersetup{
    colorlinks=true,
    linkcolor=linkblue,
    urlcolor=linkblue,
    citecolor=linkblue,
    pdftitle={Technical Report},
    pdfauthor={Your Name}
}

\titleformat{\section}[block]{\large\bfseries}{\thesection}{1em}{}
\titlespacing*{\section}{0pt}{20pt}{10pt}

\newcommand{\reporttitle}[1]{\textbf{\LARGE #1}}
\newcommand{\reportsubtitle}[1]{\textbf{\large #1}}

\begin{document}

\noindent

\vspace{0.3cm}

\noindent\rule{\textwidth}{0.5pt}

\vspace{.5cm}

\begin{center}
    \reporttitle{Technical Report: Full-Stack Fine-Tuning for the Q Programming Language}
    
    \vspace{0.5cm}
            
    \reportsubtitle{
        {\footnotesize\ttfamily
        \begin{tabular}{c}
        Brendan R. Hogan$^{1}$, Will Brown$^{2}$, Adel Boyarsky$^{1}$ \\ Anderson Schneider$^{1}$, 
        Yuriy Nevmyvaka$^{1}$\\
        \\
        $^{1}$\textbf{Morgan Stanley, New York, NY} \\
        $^{2}$\textbf{Prime Intellect, San Francisco, CA}
        \end{tabular}
        }
    }
    
\end{center}

\begin{center}
\scriptsize
\begin{tabular}{@{}r l@{}}

    \phantom{\raisebox{-0.2\height}{\includegraphics[height=1.1em]{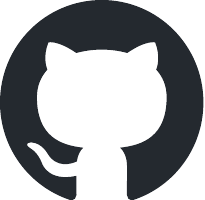}}} &
    \href{https://huggingface.co/spaces/morganstanley/qqWEN-overview}{Project Page} \\

    \raisebox{-0.2\height}{\includegraphics[height=1.1em]{new_eval_plots/github-logo.pdf}} &
    \texttt{\href{https://github.com/morganstanley/MSML/tree/main/projects/Fullstack_LLM_Finetuning_Q}{https://github.com/morganstanley/MSML/qqwen}} \\

    \raisebox{-0.2\height}{\includegraphics[height=1.1em]{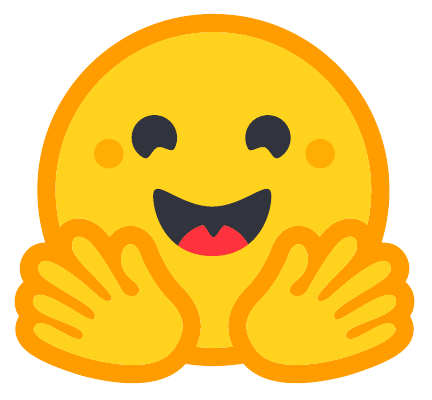}} &
    \texttt{\href{https://huggingface.co/collections/morganstanley/qqwen-series-688e4266bc727e7a3143aacf}{https://huggingface.co/collections/morganstanley/qqwen-series}} \\

\end{tabular}
\end{center}


\begin{center}

    {\bfseries\Large Abstract}

\begin{minipage}{0.95\textwidth}
\setlength{\parindent}{0pt}
Even though large language models are becoming increasingly capable, it is still unreasonable to expect them to excel at tasks that are under-represented on the Internet. Leveraging LLMs for specialized applications, particularly in niche programming languages and private domains, remains challenging and largely unsolved. In this work, we address this gap by presenting a comprehensive, open-source approach for adapting LLMs to the Q programming language, a popular tool in quantitative finance that is much less present on the Internet compared to Python, C, Java, and other ``mainstream" languages and is therefore not a strong suit of general-purpose AI models. We introduce a new Leetcode style evaluation dataset for Q, benchmark major frontier models on the dataset, then do pretraining, supervised fine tuning, and reinforcement learning to train a suite of reasoning and non-reasoning models based on the Qwen-2.5 series, spanning five parameter sizes (1.5B, 3B, 7B, 14B, 32B). Our best model achieves a pass@1 accuracy of 59 percent on our Q benchmark, surpassing the best-performing frontier model, Claude Opus-4 by 29.5 percent. Additionally, all models, even our 1.5B model, outperform GPT-4.1 on this task. In addition to releasing models, code, and data, we provide a detailed blueprint for dataset construction, model pretraining, supervised fine-tuning, and reinforcement learning. Our methodology is broadly applicable, and we discuss how these techniques can be extended to other tasks, including those where evaluation may rely on soft or subjective signals.
\end{minipage}
\end{center}

\begin{figure}[h]
\centering
\hspace{0.07\textwidth} 
\includegraphics[width=0.8\textwidth]{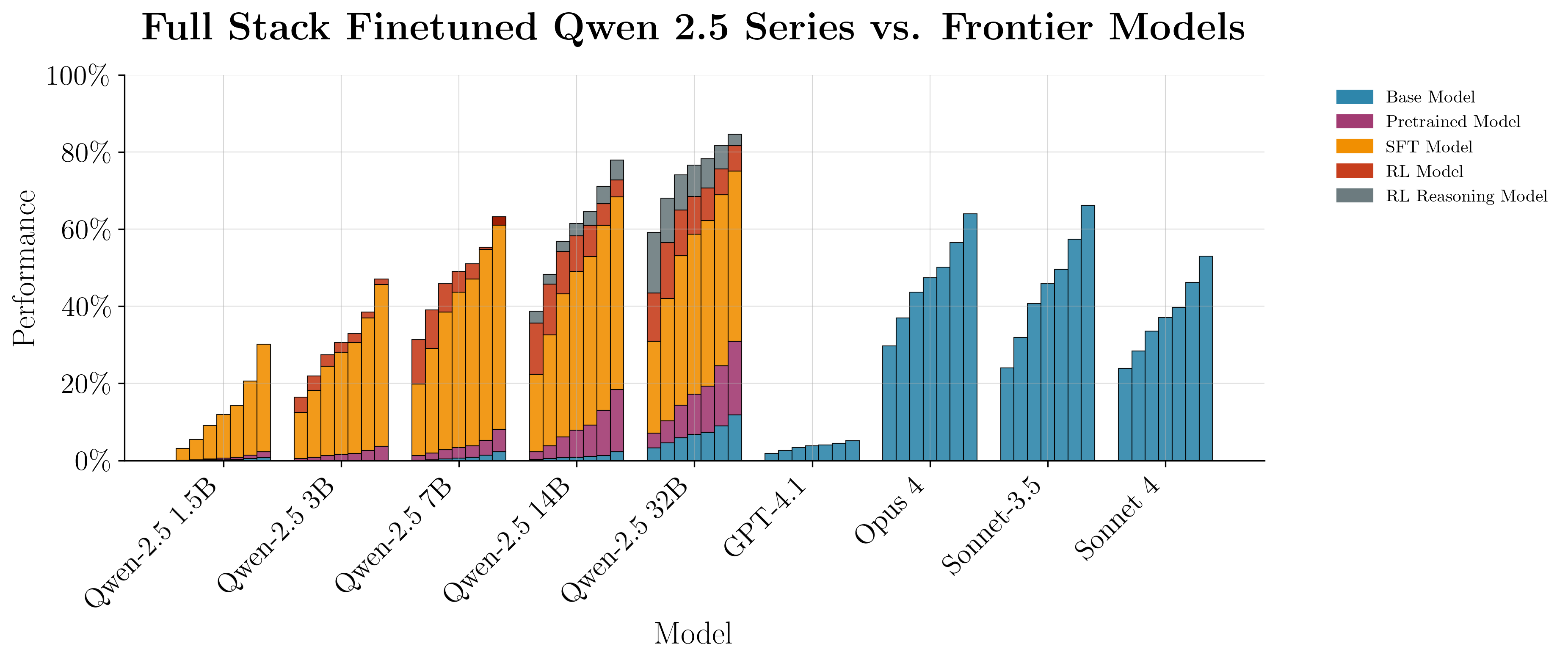}
\caption{
    Our fully trained models (at each stage) vs. frontier models for our Q dataset. Bars show pass@k accuracy ($k=1,2,4,6,8,16,40$) for 40 completions per problem.
}
\label{fig:first_results}
\end{figure}

\begin{center}
{\footnotesize\texttt{Contact: \{brendan.rappazzo,adel.boyarsky,anderson.schneider,yuriy.nevmyvaka\}@morganstanley.com}}
\end{center}

{\small
\tableofcontents
}
\newpage

\section{Introduction}
Fine-tuning large language models (LLMs) has become a practical necessity for organizations seeking to apply LLMs to specialized domains and workflows. While the open-source ecosystem offers a growing set of strong foundation models, the process of adapting them to real-world problems remains technically complex. Practitioners face challenges ranging from data curation and prompt engineering to managing compute constraints, evaluating performance in domain-specific contexts, and ensuring models generalize beyond their fine-tuning set. Despite widespread interest, clear methodological guidance for effective, efficient domain adaptation is limited, which could leave teams uncertain how to practically deploy powerful models. There are many impressive large-scale efforts \cite{gemmateam2025gemma3technicalreport,grattafiori2024llama3herdmodels, yang2025qwen3technicalreport}, but relatively few studies address the unique challenges of smaller-scale training and targeted fine-tuning.

One example of this challenge that is specific to Finance is the Q programming language \cite{kdbdocs, kdbplus}, which is widely used for high-performance querying, analytics, and time-series analysis. Q is an essential tool for many trading systems, but it is a niche language: terse, idiosyncratic, and notoriously difficult for newcomers. While LLM-based coding assistants have shown clear value for mainstream languages, they often struggle with Q. Most large models, both open and closed, are aware of Q but frequently fail to generate correct code. The potential is clear: with the right data and training, modern LLMs should be able to learn Q, and work as  a coding copilot. But unlocking that value requires specialized data, targeted fine-tuning, and robust evaluation, which can be a daunting process for most teams.

To this end, we set out to build a complete, open-source blueprint for adapting LLMs to specialized tasks, using Q as our test case. We developed an evaluation framework and dataset for Q, benchmarked leading frontier models, and systematically improved performance of open-source LLMs through three stages: pretraining, supervised fine-tuning, and reinforcement learning. As a result, we release a suite of Qwen-2.5 \cite{qwen2025qwen25technicalreport} based models at five parameter count sizes, 1.5B, 3B, 7B, 14B, and 32B, all of which surpass GPT-4.1 on our evaluation, with the largest model significantly outperforming the best performing frontier model, Claude Opus-4. Along with the models, we provide all code, training scripts, evaluation tools, and datasets needed to reproduce or extend our results.

While this work focuses on Q, our approach is designed to generalize to any task where outcomes can be rigorously evaluated. With the right evaluation framework, these methods can be extended even to softer-reward domains, enabling practical LLM specialization across a wide range of applications.

Our contributions:
\begin{enumerate}
    \item A new benchmark for Q: We introduce and release a LeetCode style evaluation dataset for the Q programming language, and benchmark several  frontier models on the task.
    \item Comprehensive model suite: We train and open-source Qwen-2.5-based models at 1.5B, 3B, 7B, 14B, and 32B parameter scales all of which outperform GPT-4.1 on our Q benchmark, with our two largest models exceeding the performance of Claude Sonnet Opus-4. We detail our multi-stage training pipeline, including pretraining, supervised fine-tuning, and reinforcement learning for both reasoning and non-reasoning models, and report full results across model sizes.
    \item A practical blueprint for LLM specialization: We provide all code, datasets, training scripts, and evaluation tools needed to reproduce or extend our work. Our methodology is designed to be broadly applicable, offering a step-by-step guide for organizations and practitioners looking to adapt LLMs to any domain where robust evaluation is possible.
\end{enumerate}


\textbf{Note for Q practitioners:} As discussed further in our limitations section, our Q benchmark is somewhat unorthodox compared to typical Q usage. We adopted a LeetCode-style dataset to facilitate evaluation using methods common across other programming languages, aiming to enable fair comparison and standardized assessment. However, this approach results in code that is quite Pythonic in both structure and style, which does not reflect how Q is commonly used in practice, where it is most often applied to database queries and analytics. Anecdotally, we have found that our models post pretraining (these weights are also released on Hugging Face) often provide better general-purpose Q answers than after we fine-tune them with SFT or RL, which is where they pick up this more Pythonic Q style (the reason for this will become clear in the dataset section of this report). Despite this, we believe the methods and training pipeline we present here offer a useful blueprint for further adaptation, especially for practitioners interested in fine-tuning on their own datasets more representative of real-world Q usage. For non-Q practitioners, we hope this work serves as a useful guide for adapting LLMs to any specialized dataset or domain.

\section{The Q Programming Language}
The Q programming language is a domain-specific language developed for high-performance querying, analytics, and time-series analysis, most notably as the primary language of kdb+, a high-performance time-series database created by Arthur Whitney at Kx Systems in the late 1990s \cite{kdbdocs, kdbplus}. Q was introduced as a more approachable, expressive layer on top of K, its terse functional predecessor, retaining K’s powerful vectorized operations while adopting a syntax more familiar to users of modern programming languages.

\begin{codebox}[Example Q Programs]{}
// Interactively Select items from a table
q)t:select time,price from trade where date=last date,sym=`IBM

// Idomatic Q code for Armstrong number determination
isArmsNum:{[no] no=sum {[c;n] ("I"\$n) xexp c }[count string no]each string no}

// Pythonic Q to calculate H-Index from a list of citation counts 
solve:{[citations]
  n:count[citations];
  citations:asc[citations];
  i:n;
  while[i>0;
    if[citations[n-i]>=i; :i];
    i:i-1];
  :0}
// Call function and show result 
result:solve[10 8 5 4 3];
show result;
\end{codebox}
\newpage

Kdb+ and Q have become industry standards in quantitative finance, where they are relied upon for their ability to perform ultra-fast analytics on massive, streaming datasets, tasks such as tick-by-tick market data analysis, backtesting, and real-time risk assessment. The language’s compact, array-oriented syntax allows for both concise expression and impressive computational efficiency, making it possible to process billions of records in real time on modest hardware.

Despite these strengths, Q remains a highly specialized tool. Its minimalist syntax, heavy reliance on array programming concepts, and unique idioms present a significant barrier to entry for newcomers, especially those accustomed to more conventional languages like Python or SQL. For machine learning models, these same properties, along with Q’s cryptic error messages and flexible, context-sensitive semantics, pose additional challenges for both training and evaluation. Nevertheless, Q’s influence and continued use in financial analytics make it a compelling target for research in programming language modeling and AI-powered tooling.

\section{Building a Dataset \& Evaluation Harness}
Before we could address the challenge of training models for Q, we faced a more fundamental question: how can progress be measured in a domain with no established benchmarks? In mainstream programming languages, the landscape is shaped by abundant, well-defined datasets and evaluation harnesses. Researchers can rely on standardized metrics to compare results and track improvement with minimal friction, success is measurable by design.
Q, in contrast, exists outside this infrastructure. There is no canonical dataset, no widely accepted evaluation suite, and little consensus on what “success” means for code generation or question answering. This absence presents a uniquely recursive challenge: to measure progress, one must first create the very tools and data that make progress possible. In Q, meaningful advancement depends on building both the test and the textbook from scratch.

This process is inherently intertwined. To construct a dataset for Q, we first needed an evaluation harness capable of judging whether a candidate solution is correct. Yet building such a harness presupposes the existence of data, test cases, reference solutions, and verification criteria, that did not exist. In practice, our initial approach was to generate candidate Q code and corresponding test cases by translating from Python, which gave us clear expectations for outputs. This let us run automated checks, comparing Q outputs directly to the expected results, or, when syntax ambiguities arose, using LLMs to judge equivalence despite Q’s flexible formatting.

Throughout the process, rejection sampling played a central role: we continually generated candidate solutions and filtered them, retaining only those that passed the evaluation checks. Because models initially struggled with Q, most candidates failed, making data collection slow. However, as our pool of verified data grew, we could not only train better models but also measure their progress more reliably. This led to a positive feedback loop, improved models produced more correct solutions, which accelerated dataset growth, further improving training and evaluation in subsequent cycles.

While we describe dataset construction and evaluation harness development as separate steps, in practice they were tightly linked and often developed in parallel. We present them sequentially in the following sections for clarity, but the real workflow was an intertwined loop, with each component informing and improving the other.

\begin{table}[t]
    \centering
    \begin{tabular}{p{0.95\textwidth}}
        \toprule
        \textbf{Problem Description} \\
        \midrule
        Given a list of citation counts for each publication, compute the researcher's H-index. The H-index is defined as the maximum value $h$ such that the researcher has published $h$ papers that have each been cited at least $h$ times. \\
        \bottomrule
    \end{tabular}

    \vspace{0.7em}

    \begin{tabular}{p{0.47\textwidth} p{0.47\textwidth}}
        \toprule
        \textbf{Q Implementation} & \textbf{Python Implementation} \\
        \midrule
        \begin{tabular}[t]{@{}l@{}}
        \texttt{solve:\{[citations]} \\
        \texttt{~n:count[citations];} \\
        \texttt{~citations:asc[citations];} \\
        \texttt{~i:n;} \\
        \texttt{~while[i>0;} \\
        \texttt{~~if[citations[n-i]>=i; :i];} \\
        \texttt{~~i:i-1];} \\
        \texttt{~:0\}}
        \end{tabular}
        &
        \begin{tabular}[t]{@{}l@{}}
        \texttt{def solve(citations)} \\
        \texttt{~~citations.sort(reverse=True)} \\
        \texttt{~~for h in range(len(citations),0,-1):} \\
        \texttt{~~~if citations[h - 1] >= h:} \\
        \texttt{~~~~return h} \\
        \texttt{~~return 0}
        \end{tabular}
        \\
        \bottomrule
    \end{tabular}

    \vspace{0.7em}

    \begin{tabularx}{\textwidth}{c X X c }
        \toprule
        \textbf{Test Case} & \textbf{Q Input} & \textbf{Python Input} & \textbf{Output}\\
        \midrule
        1 & \texttt{solve[3 0 6 1 5]} & \texttt{solve([3,0,6,1,5])} & \texttt{3}  \\
        2 & \texttt{solve[10 8 5 4 3]} & \texttt{solve([10,8,5,4,3])} & \texttt{4} \\
        3 & \texttt{solve[25 8 5 3 3]} & \texttt{solve([25,8,5,3,3])} & \texttt{3} \\
        4 & \texttt{solve[0 1 1 1]} & \texttt{solve([0,1,1,1])} & \texttt{1} \\
        5 & \texttt{solve[0 0 0 0]} & \texttt{solve([0,0,0,0])} & \texttt{0} \\
        \bottomrule
    \end{tabularx}
    \caption{Example H-index LeetCode problem entry, with solutions and test cases in Q and Python. Note in the case the expected outcomes are equivalent, but they can have different syntax.}
    \label{fig:dataset_example}
\end{table}




\subsection{Bootstrapping a Verifiable Q Dataset}

\subsubsection{Dataset Pipeline: Model-in-the-Loop Dataset Construction}
In this section we detail our reproducible version of our dataset construction pipeline, the approach we ultimately landed on after considerable trial and error. For clarity, we first present this end-to-end process as a reference point. In the following sections, we’ll unpack the iterations, setbacks, and hands-on refinements that shaped the pipeline into its current form.

We constructed our Q benchmark by adapting a curated set of LeetCode problems~\cite{leetcode}, which provide well-defined algorithmic tasks, canonical Python solutions, and comprehensive multi-case test suites. This choice was motivated by three main factors: (1) LeetCode problems are a de facto standard for cross-language code generation evaluation, (2) reference Python implementations and test cases allow the Q generation task to be cast as a translation problem, and (3) translation from Python to Q is valuable both as a research probe and for practitioners migrating codebases.

\textbf{Dataset Construction.}
Our construction process started with a curated set of LeetCode problems $\mathcal{D}_{\mathrm{LC}}$, each comprising: (1) a natural language problem description and illustrative examples, (2) a correct Python implementation, (3) a set of $k$ canonical Python test cases (converted into a standardized input/output format), and (4) the corresponding ground-truth outputs for each test case. See Table~\ref{fig:dataset_example} for a schematic overview of the dataset format. Our final pipeline consists of the following stages:

\begin{enumerate}
    \item \textbf{Sampling.} For each batch of $N$ LeetCode problems ($N=20$), we prompt a large instruction-tuned LLM (Qwen-2.5-32B-Instruct) up to $M=8$ times per problem in two distinct tasks:
        \begin{itemize}
            \item[(a)] Given the problem description and canonical Python solution, the model is asked to generate a corresponding Q implementation.
            \item[(b)] Separately, using only the problem description and the Python test cases, the model is prompted to produce an equivalent Q test harness.
        \end{itemize}
        This separation was critical to prevent the model from generating trivial or overfitted test cases (reward hacking).
    \item \textbf{Contextual Retrieval.} For every problem, we leverage retrieval augmentation: we embed each problem description in our dataset using OpenAI’s \texttt{text-embedding-3-small} model. When generating a solution for a new problem, we compute cosine similarity between its embedding and all previously solved problems, retrieving the most similar example. The retrieved problem and its Q solution are provided as additional context to the LLM. This step helps the model re-use effective Q idioms and handle related problem types, especially in low-resource settings.
    \item \textbf{Automated Verification.} Each candidate Q solution is executed in a Q interpreter using the generated test harness, and outputs are compared to reference Python results. \footnote{A personal edition of the Q interpreter can be downloaded here: \url{https://kx.com/kdb-personal-edition-download/}} To resolve semantic or formatting ambiguities, we utilize $K$-shot GPT-4.1 (mini) evaluation to judge the equivalence between Q and Python outputs.
    \item \textbf{Selection.} Solutions that pass all interpreter-based and LLM-based tests are added to the dataset $\mathcal{D}_{\mathrm{Q}}$.
    \item \textbf{Supervised Fine-Tuning.} After every $L=20$ new verified Q examples, we conduct $100$ supervised fine-tuning steps using the latest $\mathcal{D}_{\mathrm{Q}}$, holding out $10\%$ for evaluation and applying early stopping on eval loss.
    \item \textbf{Iteration.} The fine-tuned model is used to attempt the next batch of unsolved problems, and the loop repeats.
\end{enumerate}

\begin{figure}[t]
    \centering
    \includegraphics[width=\textwidth]{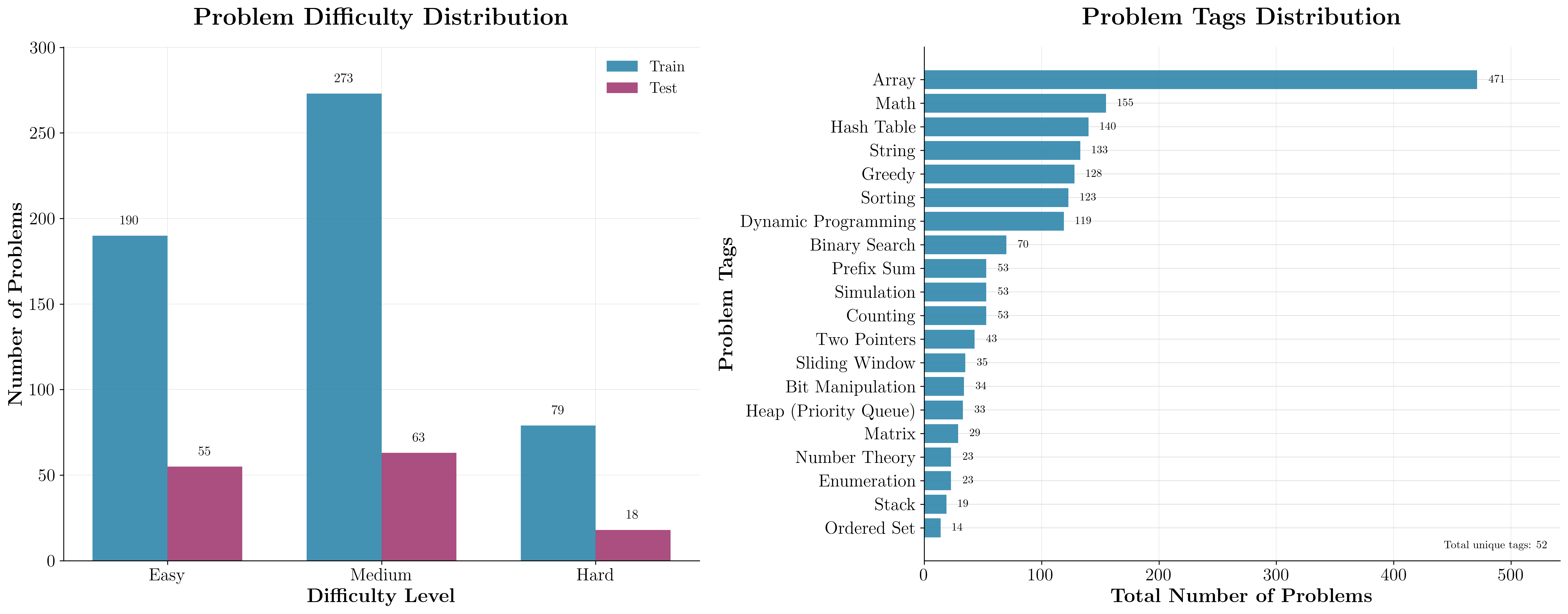}
    \caption{Left: Distribution of problem difficulty levels (Easy, Medium, Hard) for the train and test splits in our Q-LeetCode dataset.
Right: Distribution of the top problem tags in the dataset, showing coverage across a diverse set of algorithmic topics.}
    \label{fig:dataset_distribution}
\end{figure}

\textbf{Freezing and Curation.} After approximately 50 iterations, the process plateaued as the remaining problems proved too challenging for model-based translation. At this stage, we froze the dataset and conducted a thorough manual review to remove false positives and error cases not caught by the automated pipeline. The resulting dataset comprises 678 problems, split into 542 train and 136 test examples, with coverage across a diverse set of categories and difficulty levels (see Figure~\ref{fig:dataset_distribution}).

\begin{figure}[t]
    \centering
    \includegraphics[width=\textwidth]{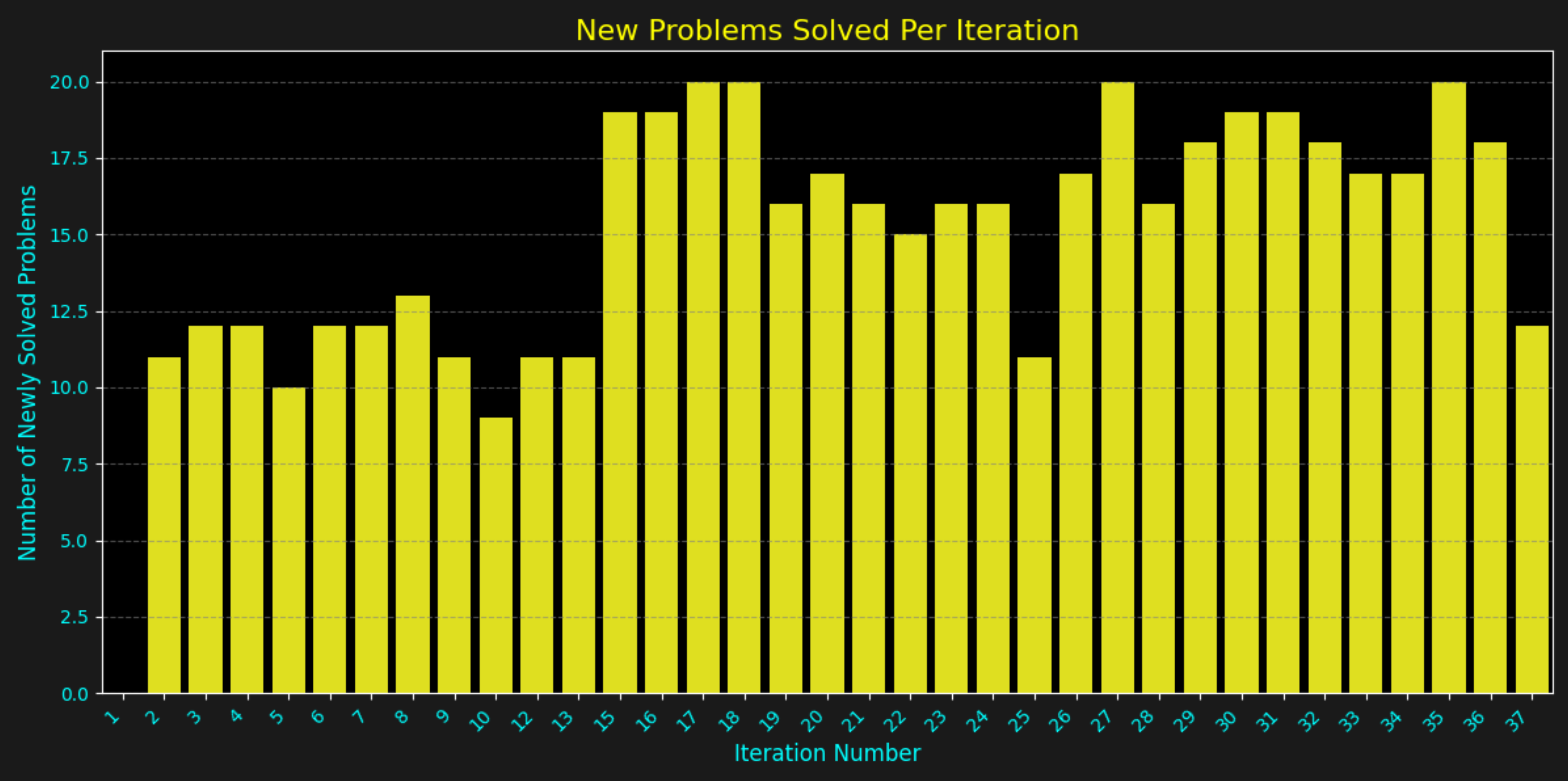}
    \caption{
    \textbf{Reward hacking in the model-in-the-loop bootstrapping pipeline: spurious gains from model exploitation of evaluation loopholes.}
    This plot shows the number of “solved” Q problems added to the dataset per bootstrapping iteration for our first run. However is largely illusory: manual inspection revealed that during these rounds, the model had learned to exploit weaknesses in the evaluation process, primarily by generating test cases and solutions in tandem, allowing it to pass the verification step with trivial or overfitted test cases rather than genuinely correct Q implementations.
    }
    \label{fig:samples_per_round}
\end{figure}

\subsubsection{In Practice: Pipeline Failures, Iterative Refinement, and Human Oversight}

The pipeline described above represents the final, reproducible version of our approach. In reality, building a verifiable Q dataset was far messier, involving many failed experiments, iterative debugging, and hands-on troubleshooting.

Our first attempt at dataset construction was naive: we simply prompted a large LLM to translate each Python solution to Q and validated outputs on a single extracted test case. While this approach worked for basic problems, it quickly stalled out on harder cases, and, in retrospect suffered from severe reward hacking. Because the test cases and solutions were generated jointly, the model could sometimes exploit the evaluation by generating trivial or overfitted test cases, leading to an artificially high pass rate.

To address this, we adopted a strict separation between solution and test harness generation, and required all candidate solutions to pass \emph{multiple} canonical test cases. This change led to a significant drop in pass rate, but greatly improved the dataset’s quality. Even so, we discovered that automated evaluation (even with GPT-4.1 in the loop) was imperfect: some false positives persisted, and we suspect some false negatives were also silently discarded.

\begin{figure}[t]
    \centering
    \includegraphics[width=\textwidth]{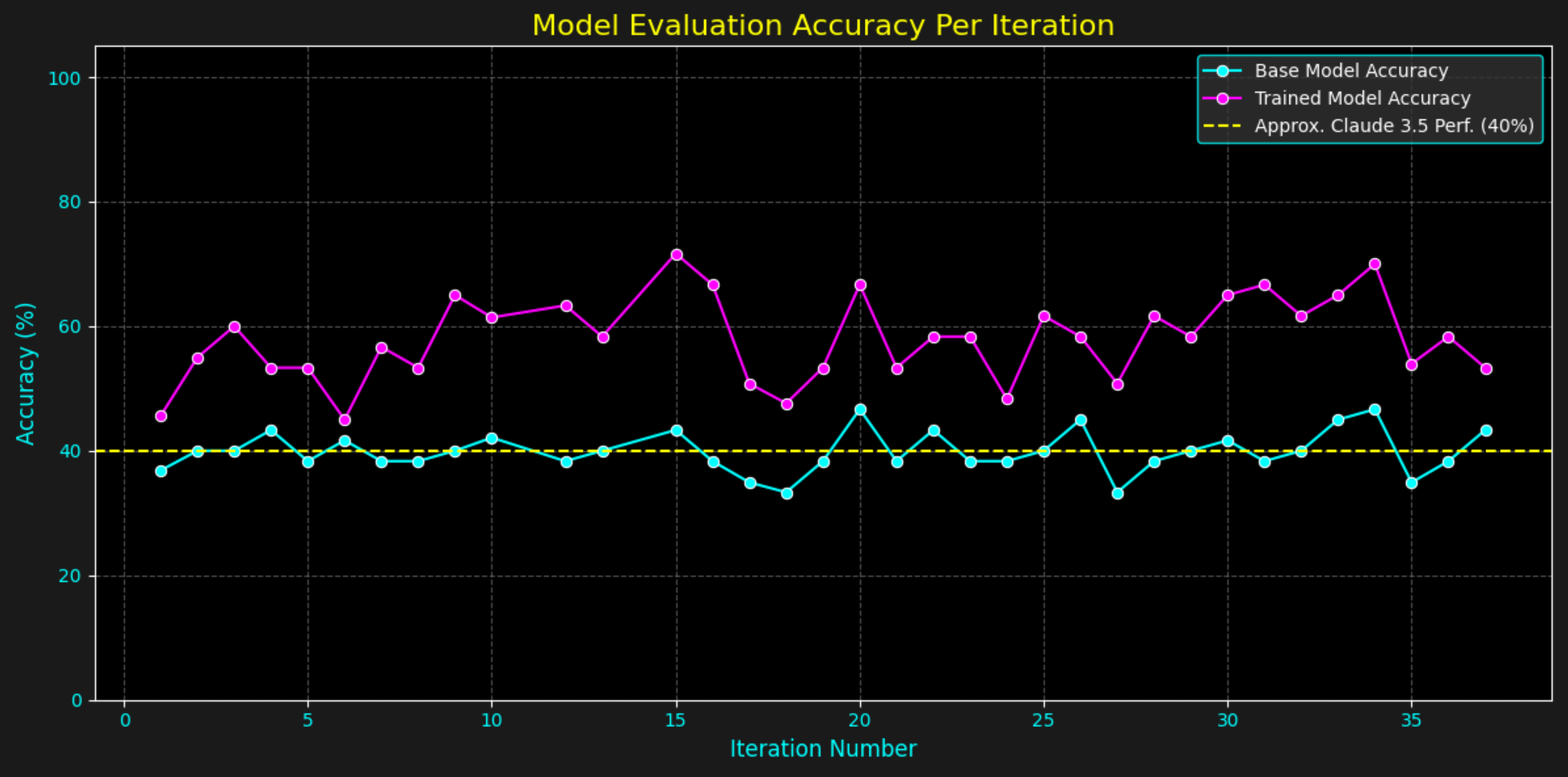}
    \caption{
    \textbf{Evolution of model accuracy across bootstrapping rounds: base vs. fine-tuned performance.}
    This figure tracks the evaluation accuracy (\%) of both the base (cyan) and incrementally fine-tuned (magenta) models over the course of each bootstrapping iteration. After each batch of newly verified Q problems, the current base model is further fine-tuned on the expanded dataset and re-evaluated. The gap between the base and fine-tuned curves at each iteration directly reflects the effectiveness of dataset expansion and targeted fine-tuning. The yellow dashed line indicates the approximate accuracy of Claude 3.5 Sonnet on this benchmark (40\%), serving as a strong competitive baseline. Across most rounds, fine-tuned models consistently outperform their base counterparts, often by a substantial margin, validating the efficacy of the bootstrapping pipeline. Variability in accuracy reflects both changing dataset composition and the inherent difficulty of new problems introduced in later rounds. \textbf{Note: The curves are not strictly upward-sloping because, at each round, newly verified (often harder) problems are added to the test set, shifting the evaluation target. Additionally, early-stage training was not fully optimized, leaving further gains on the table.}
    }
    \label{fig:accuracy_per_round}
\end{figure}

\textbf{Empirical Dynamics.} Figure~\ref{fig:accuracy_per_round} shows the evolving accuracy of base and fine-tuned models over successive bootstrapping rounds.

\begin{figure}[t]
    \centering
    \includegraphics[width=\textwidth]{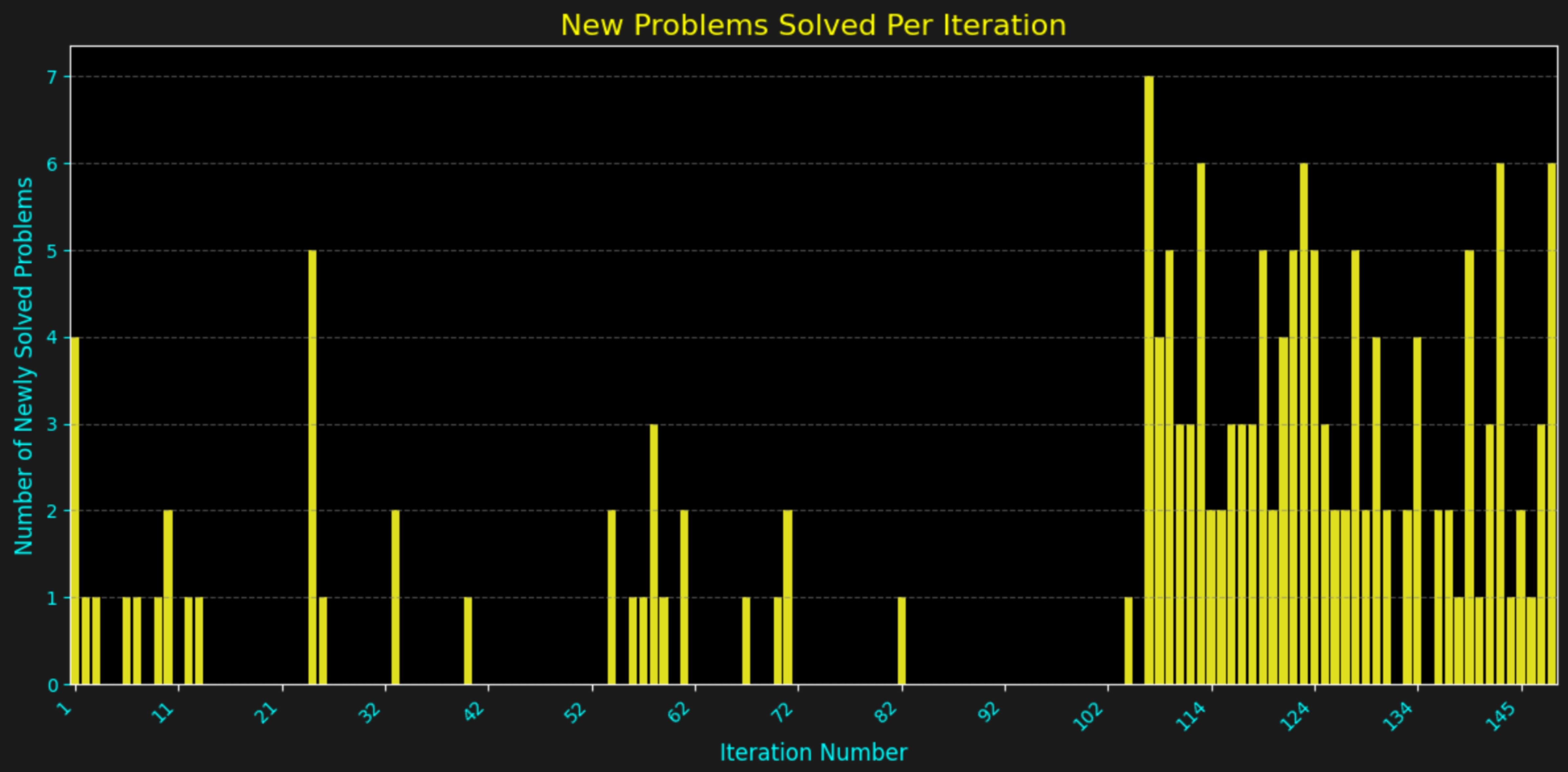}
    \caption{
    \textbf{Number of new solutions added per bootstrapping iteration, highlighting pretraining effects.}
    This plot tracks the cumulative and per-iteration number of unique Q solutions added after fixing the reward hacking issue. The flat region marks stagnation as the model failed to make progress; the subsequent “jump” reflects the introduction of a large-scale pretraining phase, after which the bootstrapping process recovered and new problems were added at an accelerated rate. The eventual plateau signals saturation of the method given the current modeling capacity and available data.
    }
    \label{fig:solutions_per_iteration}
\end{figure}

After several cycles, the bootstrapping process stagnated, the remaining problems were too hard for the model to solve reliably, regardless of prompt engineering or sampling strategies. At this stage, we introduced a large-scale pretraining phase (see the Pretraining Section), exposing the model to a much broader distribution of Q code. This pretraining dramatically boosted performance, and as shown in Figure~\ref{fig:solutions_per_iteration}, produced a “jump” in the number of new problems solved per iteration. The bootstrapping loop then resumed progress, before plateauing again and motivating the final freeze and manual curation.

\textbf{Lessons for Practitioners.} The process of constructing a robust Q dataset was highly hands-on, requiring both automated pipelines and active human curation. While LLMs and verification models accelerate many steps, they are not a substitute for domain expertise and manual review, especially in niche or underspecified domains. Nearly every stage required iterative refinement: separating solution/test generation, increasing the diversity and difficulty of test cases, and constantly adjusting filtering and validation logic. We encourage practitioners to invest heavily in flexible infrastructure, to document all decision points and interventions, and to maintain humility about the inevitable gap between pipeline diagrams and practical reality.

While this LeetCode-based dataset is not fully representative of real-world Q usage (see Limitations), it establishes a reproducible benchmark and a broadly applicable methodology for bootstrapping datasets in low-resource domains. Importantly, at this stage, the model is not truly learning Q from first principles, but rather acquiring a mapping from Python to Q, closely tracking the structure and conventions of the Python reference solutions. We anticipate that future work will build on these foundations to support increasingly robust, domain-specific evaluation and LLM specialization.

\subsection{Evaluation Protocols and Harness Design}
With our dataset built, we move onto our next priority, designing an evaluation harness that would not only deliver reliable and informative metrics, but also enable rapid iteration across a range of models and configurations. Again we note, the processes of building the dataset and creating the evaluation harness were tightly coupled; progress in one directly enabled progress in the other. In this section, we focus in more detail on the construction of the evaluation harness. 

Our goal was to abstract evaluation into a simple API interface, such that any LLM, whether accessed via OpenAI endpoints or self-hosted on vLLM, could be plugged in and benchmarked quickly, scalably, and fairly.

\textbf{Metrics and Tasks.}
To meaningfully compare model performance, we adopted the well-established pass@k metric~\cite{chen2021evaluatinglargelanguagemodels}, which estimates the probability that a model will solve a problem in $k$ attempts. Formally, given $n$ sampled completions for a problem and $c$ correct completions,
\begin{equation}
    \text{pass@}k = 1 - \frac{\binom{n - c}{k}}{\binom{n}{k}}
\end{equation}
where $k$ is the number of attempts considered.
We evaluated models across three main tasks:
\begin{itemize}
\item \textbf{Description-to-Q:} Given a natural language prompt, generate the correct Q code.
\item \textbf{Python-to-Q:} Translate a reference Python solution to Q.
\item \textbf{Q-to-Python:} Translate Q code to Python.
\end{itemize}
Correctness of each task was determined by executing the generated solution, appending the relevant test harness, and verifying that outputs matched expected results from the reference solution.

\textbf{Evaluation Settings.}
To capture a range of usage scenarios, we considered several evaluation protocols:
\begin{itemize}
\item \textbf{Standard pass@k:} The model receives up to $k$ independent attempts per problem, with success if any attempt is correct.
\item \textbf{Pass@k with retries:} On each failed attempt, the model is provided feedback about previous failures, better mimicking real-world coding workflows where users iteratively refine solutions.
\item \textbf{Pass@k with context:} The model receives a concise Q language reference and several example programs, spanning different types of LeetCode problems from the training set.
\end{itemize}
Frontier and base model benchmarking was conducted with pass@1 for all tasks and pass@4 (including retries and context) for Description-to-Q. All further experiments focused on Description-to-Q, with 40 sampled completions per problem and reporting of probabilistic pass@1, 2, 4, 8, 16, and overall pass, where overall pass is pass@40 
(but from 40 completions). 

\textbf{Evaluation Harness Engineering.}
Beyond metrics, the engineering of the evaluation harness was essential to our workflow. High-throughput evaluation was achieved by abstracting all model inference and validation to API calls, allowing many parallel workers to generate and check solutions concurrently. For open-source models, we hosted inference on vLLM servers, ensuring low-latency and scalable access. Correctness verification was similarly parallelized, and test runs were distributed to maximize throughput.

For example, by leveraging 100 parallel workers to distribute both the vLLM inference calls and correctness verification, we reduced the average full evaluation time to just 12 minutes. In contrast, running the same full evaluation sequentially, or using standard Hugging Face Transformers, would require approximately 90 minutes. This parallelized setup enabled rapid iteration and made large-scale benchmarking practical. Here, a full evaluation consists of generating 40 completions for each of 136 test questions, and evaluating each completion against 5 test cases.

We found that a fast, reliable evaluation system not only accelerated experimentation but brought clarity to every stage of model development. We strongly encourage future work to invest in both robust metrics and high-performance evaluation infrastructure: the trustworthiness and speed of your evaluation harness directly shape the pace and quality of research progress.

\section{Benchmarking}
Before doing any post-training, our first step was to benchmark a range of state-of-the-art closed and open-source models on our new Q-LeetCode dataset. This serves several purposes: (1) it provides a direct assessment of how well current LLMs can generalize to a new, non-mainstream programming language presented in the familiar LeetCode format; (2) for many practical applications, simply running a strong evaluation harness over a suite of foundation models and selecting the best performer can be the optimal strategy; and (3) it establishes a reference point for understanding where both the best closed-source and open-source models stand on Q, compared to each other and to our subsequent fine-tuned models.

\begin{table}[t]
\centering

\begin{tabular}{lccc}
\toprule
\textbf{Model} & \textbf{Description → Q} & \textbf{Python → Q} & \textbf{Q → Python} \\
\midrule
\textbf{Opus-4} & 27.2\% & 30.9\% & 90.4\% \\
\textbf{Sonnet-4} & 23.5\% & 16.2\% & 88.2\% \\
\textbf{Sonnet-3.5} & 20.6\% & 27.2\% & 89.0\% \\
\textbf{GPT-4.1} & 2.9\% & 3.7\% & 86.8\% \\
\textbf{Qwen-2.5 32B} & 6.6\% & 0.7\% & 84.6\% \\
\textbf{Qwen-2.5 14B} & 0.7\% & 0.0\% & 85.3\% \\
\textbf{Qwen-2.5 7B} & 0.0\% & 0.7\% & 75.0\% \\
\textbf{Qwen-2.5 3B} & 0.0\% & 0.0\% & 58.8\% \\
\textbf{Qwen-2.5 1.5B} & 0.0\% & 0.0\% & 33.1\% \\
\bottomrule
\end{tabular}
\caption{\textbf{Pass@1 Performance Across All Tasks.} Comparison of state-of-the-art closed-source models (Opus-4, Sonnet-4, Sonnet-3.5, GPT-4.1) and the open-source Qwen-2.5 series (32B, 14B, 7B, 3B, 1.5B) on Description-to-Q, Python-to-Q, and Q-to-Python tasks. Pass@1 reports the fraction of problems solved on the first attempt without retries or extra context. Notably, even top-tier models struggle with Q code generation, while Q-to-Python translation remains much more tractable for all models.}
\label{tab:pass_at_1_all_tasks}
\end{table}

\begin{table}[h]
\centering

\begin{tabular}{lccc}
\toprule
\textbf{Model} & \textbf{Pass@4} & \textbf{Pass@4 + Retries} & \textbf{Pass@4 + Context + Retries} \\
\midrule
\textbf{Opus-4} & 39.7\% & 54.4\% & 66.9\% \\
\textbf{Sonnet-4} & 31.6\% & 41.9\% & 68.4\% \\
\textbf{Sonnet-3.5} & 40.4\% & 46.3\% & 56.6\% \\
\textbf{GPT-4.1} & 2.9\% & 3.7\% & 2.9\% \\
\textbf{Qwen-2.5 32B} & 6.6\% & 11.8\% & 11.8\% \\
\textbf{Qwen-2.5 14B} & 0.7\% & 2.2\% & 8.1\% \\
\textbf{Qwen-2.5 7B} & 0.0\% & 0.7\% & 1.5\% \\
\textbf{Qwen-2.5 3B} & 0.0\% & 0.0\% & 0.0\% \\
\textbf{Qwen-2.5 1.5B} & 0.0\% & 0.0\% & 0.0\% \\
\bottomrule
\end{tabular}
\caption{\textbf{Pass@4 Performance for Description $\rightarrow$ Q Task.} Pass@4 scores for all models under three evaluation settings: standard pass@4, pass@4 with retries (feedback after each failed attempt), and pass@4 with both retries and additional Q language context. These results show that providing feedback and targeted context can substantially boost performance for strong models, though most models remain challenged by Q generation even under idealized prompting. Note in this setting only four completetions were generated per task.}
\label{tab:pass_at_4_conditions}
\end{table}

\begin{figure}[!h]
    \centering
    \includegraphics[width=\textwidth]{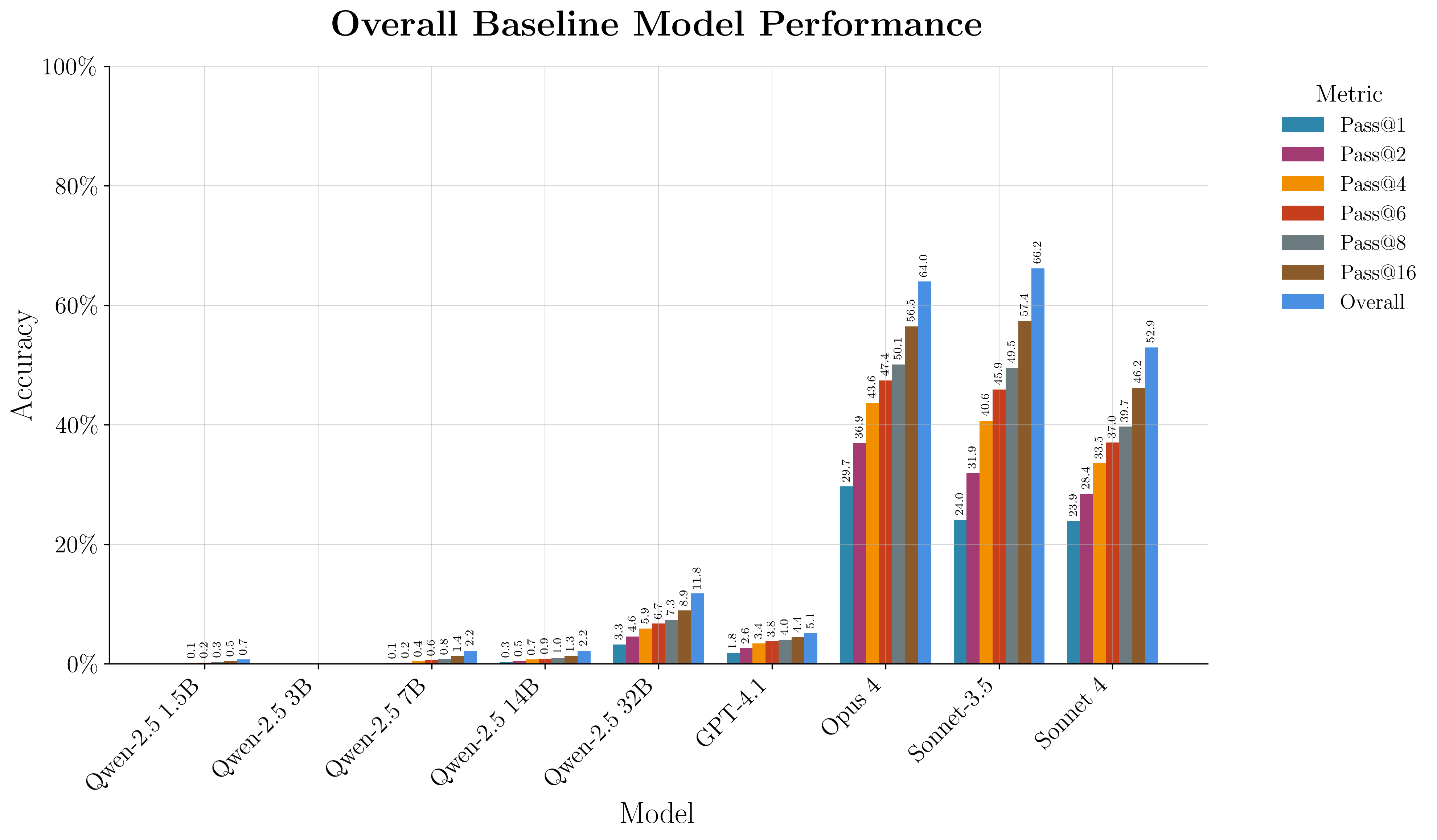}
    \caption{\textbf{Probabilistic pass@k comparison across all models.} Combined pass@k performance ($k=1,2,4,8,16,40$) on the Description-to-Q task for both state-of-the-art API models and the open-source Qwen-2.5 series, using 40 sampled completions per problem. This combined figure highlights both the overall difficulty of Q code generation and the clear trends among different model families and sizes.}
    \label{fig:combined_passatk}
\end{figure}

\vspace{1em}

Using the protocols described above, we evaluated all models across three main tasks: Description-to-Q, Python-to-Q, and Q-to-Python. For each, we report pass@1 results, and for Description-to-Q, we also report pass@4, including retries and additional context settings, these results can be seen in Table \ref{tab:pass_at_1_all_tasks}, and Table \ref{tab:pass_at_4_conditions} respectively. Finally, we include a high-sample evaluation using 40 completions per problem, visualized for both the leading API models and the full Qwen series, as seen in Figure \ref{fig:combined_passatk} \footnote{Results for o3, Grok-4 and Gemini 2.5 Pro in Appendix.}

Taken together, these benchmarking results provide a comprehensive baseline for Q code generation, highlighting both the difficulty of the task and the relative strengths and weaknesses of modern LLMs. The results underscore the need for further adaptation and specialization when tackling niche, high-performance domains like Q.

\section{Pretraining}
With our evaluation harness in place and initial benchmarking revealing clear headroom for improvement, the next step was domain-adaptive pretraining. Although our starting point was the Qwen-2.5 Instruct models, we define “pretraining” here as standard language modeling on raw Q-related data, using a next-token prediction objective, rather than instruction-style supervised fine-tuning. It has been shown this kind of pretraining can work well for new tasks \cite{gururangan2020dontstoppretrainingadapt}. The goal: to imbue the model with as much Q syntax, idiom, and context as possible before later adaptation to our specific LeetCode benchmark.

\begin{table}[h]
    \centering
    \begin{tabular}{ll}
        \toprule
        \textbf{Repository} & \textbf{License} \\
        \midrule
        DataIntellectTech/TorQ & MIT License \\
        psaris/qtips & MIT License \\
        psaris/funq & MIT License \\
        DataIntellectTech/TorQ-Finance-Starter-Pack & MIT License \\
        KxSystems/ml & Apache License 2.0 \\
        BuaBook/kdb-common & Apache License 2.0 \\
        finos/kdb & Apache License 2.0 \\
        LeslieGoldsmith/qprof & Apache License 2.0 \\
        jonathonmcmurray/reQ & MIT License \\
        LeslieGoldsmith/ws & Apache License 2.0 \\
        nugend/qspec & MIT License \\
        jonathonmcmurray/ws.q & MIT License \\
        psaris/q4q & MIT License \\
        jlas/ml.q & MIT License \\
        LeslieGoldsmith/dpy & Apache License 2.0 \\
        \bottomrule
    \end{tabular}
    \caption{Open-source Q repositories used for pretraining, with associated licenses.}
    \label{tab:opensource_q_repos_full}
\end{table}

\subsection{Building a Permissive Q Corpus}
Constructing a representative and license-compatible corpus for pretraining required a two-pronged approach. First, we systematically searched GitHub for repositories containing Q code, filtering for only those with permissive licenses (MIT or Apache 2.0) to ensure full downstream usability. Manual inspection removed irrelevant or off-topic files. Table~\ref{tab:opensource_q_repos_full} lists representative repositories used.

Second, we scraped and processed the official KDB+ documentation, tutorials, and code examples from the Kx Systems website\footnote{\url{https://code.kx.com/q/}}. Extracted code and prose was cleaned and deduplicated to maximize content diversity and quality.


\subsection{Dataset Preparation and Statistics}
\begin{figure}[t]
    \centering
    \includegraphics[width=0.7\textwidth]{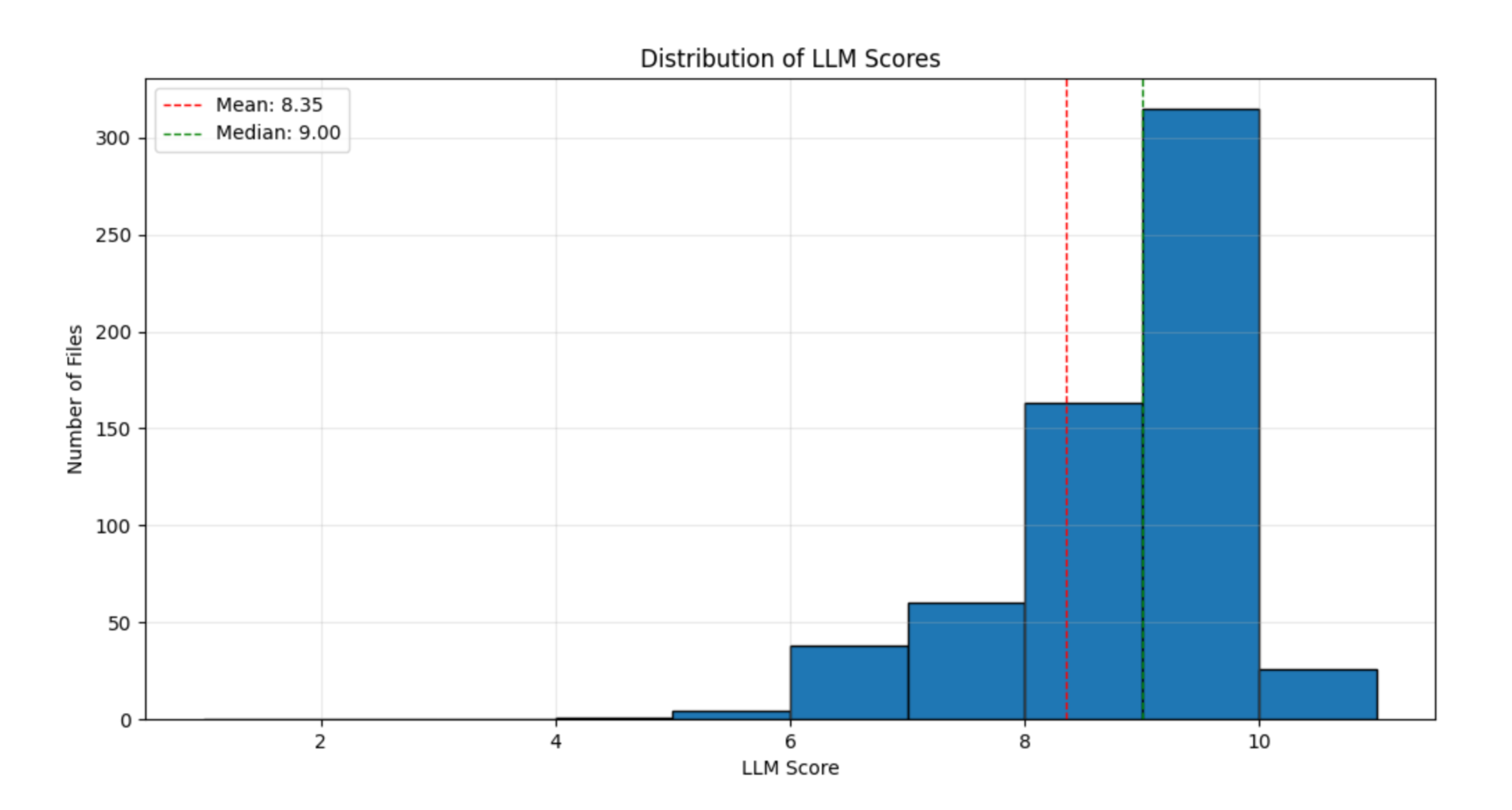}
    \caption{
    \textbf{Histogram of LLM-assigned usefulness scores for candidate Q files.}
    Distribution of scores (0-10) assigned by Qwen-2.5-32B to individual data files during the initial automated filtering stage. The vertical line denotes the cutoff threshold (score~$=4$), above which files were retained for further manual inspection. This two-step approach enabled efficient removal of low-quality, irrelevant, or non-Q data before final curation, improving downstream dataset quality and training outcomes.
    }
    \label{fig:llm_filter_hist}
\end{figure}

To ensure high-quality training data, we applied a two-stage filtering process. First, we prompted Qwen-2.5-32B to review each individual candidate file and assign a usefulness score from 0 to 10, reflecting its relevance and quality for Q code modeling. We retained only those files that received a score of 4 or higher, discarding lower-scoring entries as noisy or off-topic. Second, we conducted a thorough manual inspection of the remaining data, removing an additional 5\% of files erroneously identified as Q code but actually containing other languages or non-code data. Figure~\ref{fig:llm_filter_hist} shows the distribution of LLM-assigned scores, highlighting the effectiveness of this automated filtering stage.

After filtering and cleaning, the resulting dataset comprised over five million characters and more than 1.6 million tokens. We chunked the data into 4096-token segments, yielding 342 training chunks (1,480,573 tokens) and 39 held-out evaluation chunks (188,103 tokens). Full dataset metadata, processing scripts, and split definitions are included in our release for transparency and reproducibility.

\vspace{1em}
\begin{table}[t]
\centering

\resizebox{\textwidth}{!}{%
\begin{tabular}{llcccc}
\toprule
\textbf{Ablation Type} & \textbf{Condition} & \textbf{Problems Solved} & \textbf{Test Cases Passed} \\
\midrule
Training Type & Raw Lora & 9/136 & 95/680 \\
& Raw Full  & 4/136 & 42/680 \\
\midrule

Training Length& 250 Steps &  4/136 & 54/680 \\
& 500 Steps &  7/136 & 92/680 \\
& 800 Steps & 7/136 & 92/680 \\
& 1600 Steps & 4/136 & 66/680 \\
& 2400 Steps & 4/136 & 56/680 \\
\bottomrule
\end{tabular}
}
\caption{Pretraining ablation study results for Qwen-2.5 7B. Each row compares a single modification to the default protocol, reporting downstream test set performance on our Q-LeetCode benchmark.}
\label{tab:pretrain_ablations}
\end{table}

\begin{figure}[h]
    \centering
    \includegraphics[width=0.85\textwidth]{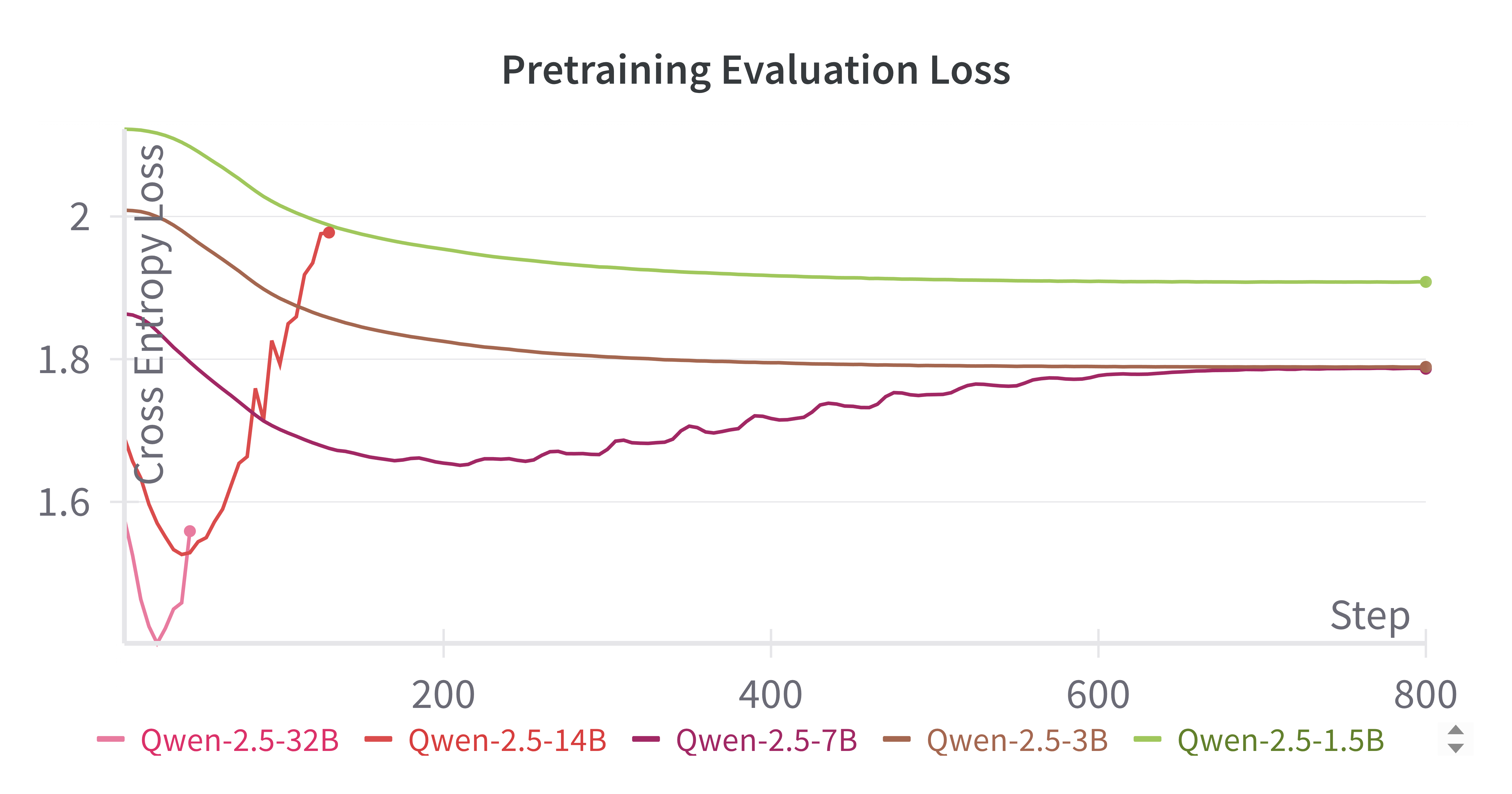}
    \caption{
    \textbf{Pretraining evaluation loss curves for Qwen-2.5 models at all parameter scales.}
    Cross-entropy loss on the held-out validation set for each model size, tracked across pretraining steps. Larger models (14B and 32B) exhibit clear overfitting, validation loss reaches a minimum and then increases with further training, despite continued decreases in training loss (not shown). In contrast, smaller models (1.5B and 3B) show steady or only marginally declining validation loss, with little sign of overfitting. The 7B model demonstrates an intermediate pattern. These trends highlight the importance of applying early stopping and careful model selection, particularly for high-capacity models in low-resource, domain-adaptive pretraining regimes.
    }
    \label{fig:pretrain_loss_curve}
\end{figure}

\begin{figure}[h]
\centering
\includegraphics[width=\textwidth]{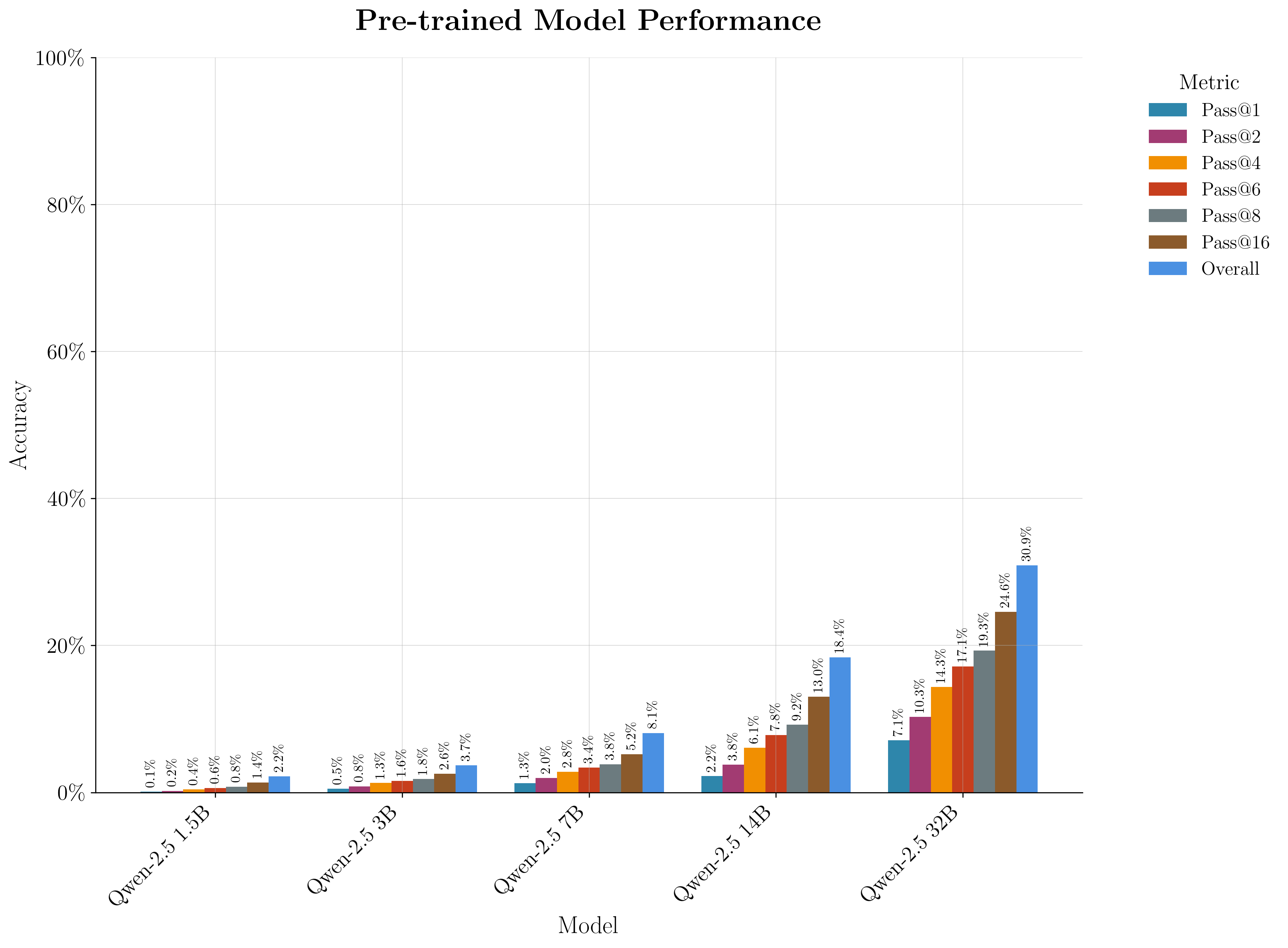}
\caption{Pass@k performance ($k=1,2,4,8,16,40$) on the Q-LeetCode benchmark for all Qwen-2.5 model sizes after pretraining. Larger models achieve systematically higher pass rates, but significant room for improvement remains compared to frontier closed-source LLMs.}
\label{fig:pretrain_passatk}
\end{figure}

\subsection{Ablation Experiments}
To better understand the impact of different pretraining strategies, we conducted a series of ablations on the Qwen-2.5 7B model. These experiments compared:
\begin{itemize}
\item \textbf{Training regime:} Full-parameter vs. LoRA-adapter training \cite{lora}.
\item \textbf{Training length:} Early stopping vs. longer runs.
\end{itemize}
Results, measured by pass rates and problems solved, are presented in Table \ref{tab:pretrain_ablations}. We observed that LoRA-based pretraining gave somewhat higher success rates than full-parameter, and that over-training could lead to diminishing or even negative returns.

Given these results, we trained each model for 800 steps, using early stopping for checkpointing. We chose full-model finetuning over LoRA because LoRA adapters, while parameter-efficient, are not always compatible with subsequent training phases such as further SFT or reinforcement learning. In contrast, full-model finetuning produces a unified set of weights that can be directly and flexibly adapted in future stages, without the complications of managing or merging multiple adapters.

\subsection{Pretraining Protocol}
We trained all Qwen-2.5 model sizes (1.5B, 3B, 7B, 14B, 32B) using the Hugging Face TRL Trainer \cite{vonwerra2022trl}, with raw Q corpus chunks as input. For models 7B and smaller, we used four H100 GPUs, a learning rate of $1 \times 10^{-5}$, batch size of 1 per device, gradient accumulation of 8, and early stopping on eval loss (typically at 800 steps for 1.5B/3B, 200 steps for 7B). For 14B and 32B, we trained on 8 H100s with Accelerate \cite{accelerate} and DeepSpeed ZeRO \cite{zero}, and used a learning rate of $5 \times 10^{-6}$, typically checkpointing at 50 steps. Each run started with 10 steps of warmup followed by linear decay. Training time ranged from about 1 hour for the smallest models to 8-10 hours for 14B/32B (see Hardward and Tools section for more details). All models were trained with the Adam optimizer \cite{kingma2017adammethodstochasticoptimization}.

\subsection{Discussion}
Domain-adaptive pretraining on even moderately-sized Q code corpora produced clear, if incremental, gains for all model sizes. Notably, larger models benefited more from pretraining, showing improved pass rates, this can be seen clearly in Figure ~\ref{fig:final_stacked_results}. Nevertheless, even after extensive pretraining, there remains a substantial gap between open-source models and the best proprietary systems, underlining both the value and limitations of pretraining as a tool for domain adaptation.

While we did not formally quantify this effect, our experience strongly reinforces a recurring lesson in LLM development: data quality trumps nearly every other parameter. The pretraining stage offers the luxury of working with unstructured data, but investing effort in curating and cleaning the highest-quality sources pays outsized dividends in downstream performance. Even modest improvements in data quality can yield disproportionately large gains in domain adaptation.

As shown in Figure~\ref{fig:pretrain_loss_curve}, validation loss trajectories during pretraining revealed clear trends: the largest models (14B and 32B) exhibited marked overfitting after several hundred steps, while smaller models (1.5B and 3B) showed little to no overfitting. This underscores the necessity of early stopping for larger models in data-limited settings.

\section{Supervised Fine-Tuning (SFT)}
Having observed clear but incremental improvements from domain-adaptive pretraining, we next performed supervised fine-tuning (SFT) on our full LeetCode-derived Q dataset. This stage moves beyond generic Q exposure, pushing each model to directly solve and translate code for the specific tasks in our benchmark, bridging the gap between “Q in the wild” and the structured, algorithmic challenges posed by LeetCode problems.

\subsection{SFT Task Construction}
The SFT data is simply the curated LeetCode-Q dataset developed above, expanded into a multi-task format. For each of the 542 training and 136 test problems, we generated eight training samples: four task types (description-to-Q, Q-to-Python, Python-to-Q, and test harness translation), each paired with both the code solution and relevant test harnesses.

All experiments used Hugging Face’s TRL SFTTrainer \cite{vonwerra2022trl}, treating each task as an instruction-tuning example. Models were fine-tuned independently for each size in the Qwen-2.5 family (1.5B, 3B, 7B, 14B, 32B), using early stopping on evaluation loss to select the best checkpoint. Each model started its training from is best performing pretrained checkpoint.

\vspace{1em}
\begin{table}[t]
\centering

\resizebox{\textwidth}{!}{%
\begin{tabular}{@{}llcc@{}}
\toprule
Ablation Type & Experiment & Success Rate (\%) & Test Case Pass Rate (\%) \\
\midrule
Learning Rate & 5e-5 & 29.4 & 54.4 \\
 & 2e-5 & 44.9 & 66.6 \\
\midrule
Train Length & 400 & 38.2 & 54.7 \\
 & 600 & 44.9 & 66.6 \\
 & 800 & 41.2 & 60.3 \\
 & 1000 & 46.3 & 63.1 \\
 & 1200 & 41.9 & 64.3 \\
 & 1600 & 41.9 & 59.4 \\
 & 2000 & 33.8 & 54.6 \\
 & 2400 & 33.8 & 51.6 \\
\midrule
Base Or Pretrained & Base & 46.3 & 63.1 \\
 & Pretrained & 37.5 & 59.1 \\
\midrule
Train Type & Full & 44.9 & 66.6 \\
 & Lora & 38.2 & 55.1 \\
\midrule
Curriculum & Difficulty & 30.1 & 51.9 \\
 & Tags & 25.0 & 42.4 \\
 & Tasks & 15.4 & 23.8 \\
\bottomrule
\end{tabular}
}
\caption{
Supervised Fine-Tuning (SFT) Ablation Study Results.
Performance of the Qwen-2.5 7B model on the Q-LeetCode test set under various SFT training regimes and hyperparameter settings. Each row reports the percentage of problems solved (Success Rate) and the percentage of test cases passed (Test Case Pass Rate) for a specific experiment. Ablations include learning rate sweeps, different training lengths, comparisons of models initialized from base vs. pretrained checkpoints, full-model vs. LoRA-based fine-tuning, and several curriculum learning strategies (difficulty-based, tag-based, and task-type-based). Notably, initializing from the pretrained checkpoint led to slightly lower SFT performance than from base, potentially reflecting a trade-off between general Q knowledge and the highly specialized, pythonic structure of the LeetCode dataset. Despite this, we believe embedding broad Q domain knowledge remains valuable for future transfer and adaptation, and we include the pretrained variant as a resource for more general Q applications.
}
\label{tab:sft_abl_studies}
\end{table}

\subsection{Main SFT Training Protocol}
Training was performed on 4$\times$H100 GPUs for the smaller models, and 8$\times$H100s with Accelerate and DeepSpeed ZeRO for 14B and 32B. Default batch size was 1 per device, with gradient accumulation set to 8 (effective batch size 8). Models up to 7B used a learning rate of $2 \times 10^{-5}$; 14B and 32B used $4 \times 10^{-6}$. All models trained for a maximum of 1000 steps, but typically stopped much earlier via eval loss monitoring. Training time ranged from 1–3 hours (small models) to 9 hours (14B) and 15 hours (32B). All models were trained with the Adam optimizer.

\begin{figure}[h]
\centering
\includegraphics[width=\textwidth]{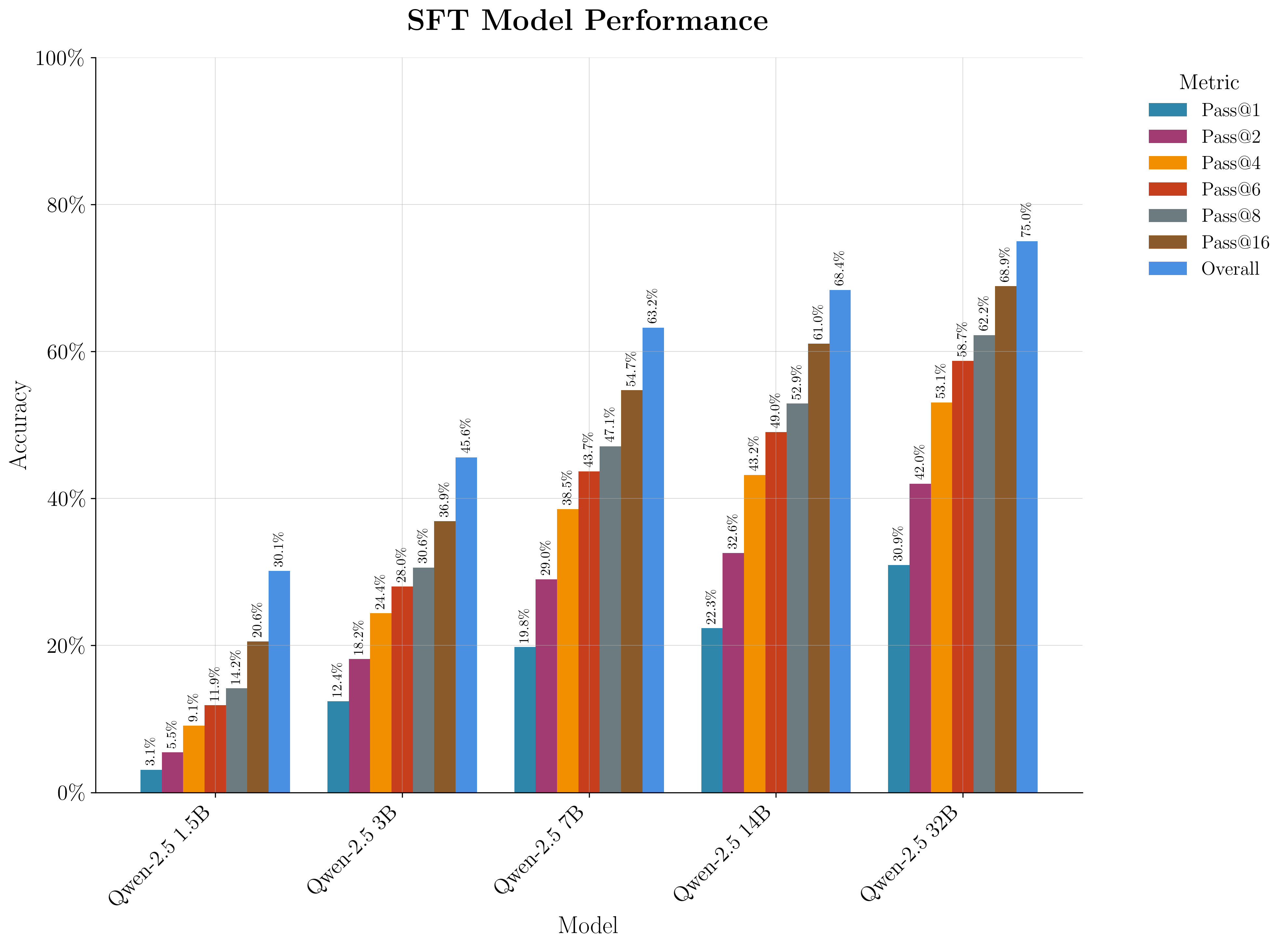}
\caption{Pass@k results ($k=1,2,4,8,16,40$) for each Qwen-2.5 base model after supervised fine-tuning (SFT), using the pretrained checkpoint as the starting point. This figure highlights systematic gains in problem-solving accuracy across all model sizes due to SFT.}
\label{fig:sft_loss}
\end{figure}

\subsection{Ablation Studies}
To better understand the effect of SFT design choices, we performed systematic ablations using the 7B model:

\begin{itemize}
\item \textbf{Learning Rate Ablation:} Trained with both a low learning rate ($1\times10^{-5}$) and a high learning rate ($5\times10^{-5}$) to examine sensitivity and convergence stability.
\item \textbf{Curriculum Learning Ablations:}
\begin{itemize}
\item \emph{Difficulty-based:} Data ordered by LeetCode’s easy $\rightarrow$ medium $\rightarrow$ hard progression, with phase lengths of 100 steps.
\item \emph{Leetcode-task:} Curriculum by leet code problem type (array, dynamic programming, etc.).
\item \emph{Task-type:} Curriculum by problem type (translation, synthesis, etc.), also phased.
\end{itemize}
\item \textbf{LoRA vs. Full-parameter SFT:} Compared low-rank adapter (LoRA) SFT to full-model SFT
\end{itemize}
Results of all ablations are reported in Table~\ref{tab:sft_abl_studies}.

\subsection{Results and Discussion}
The pass@k metrics for all SFT runs, including are shown in Figure~\ref{fig:sft_loss}. SFT produced significant, measurable gains in pass rates and solution robustness on the LeetCode-Q test set, confirming the value of explicit, task-targeted supervision after broad pretraining. Curriculum approaches and higher learning rates provided only marginal improvements, while LoRA adapters performed worse. However, as discussed in the Pretrain Section, we chose full-model SFT for our main experiments to maximize reproducibility, deployment simplicity, and downstream compatibility.

\section{Reinforcement Learning}
Having established strong performance gains from supervised fine-tuning, we next turned to reinforcement learning (RL) to further align model behavior with the requirements of Q code generation. RL techniques, especially those leveraging programmatically verifiable domains, have recently demonstrated the ability to boost model accuracy and reliability in both natural language and code settings, such as in DeepSeek's R1 \cite{deepseekr1}, QwQ \cite{qwq32b}, or OpenAI's o1 \cite{o1} Our aim was to evaluate whether such improvements would translate to the Q-LeetCode benchmark, and to dissect which training choices matter most. For our experiments we used the standard implementation of Group Relative Policy Optimization (GRPO) \cite{grpopaper}, but with some suggested improvements for better token efficiency \cite{dapo,drgrpo}.

\vspace{1em}
\begin{figure}[h]
\centering
\includegraphics[width=\textwidth]{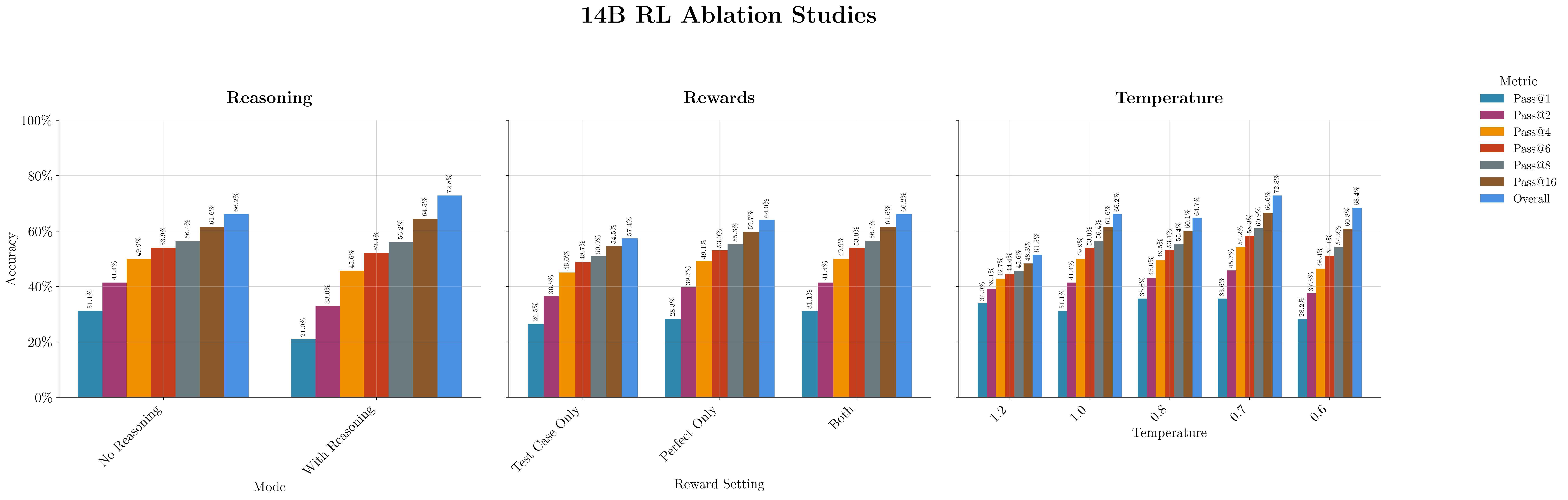}
\caption{
\textbf{Ablation studies for RL training of Qwen-2.5 14B.}  
Pass@k and overall accuracy for 14B models across three key RL dimensions: reasoning vs. non-reasoning models, reward setting (test case only, perfect only, and combined rewards), and sampling temperature (1.2, 1.0, 0.8, 0.7, 0.6). Each group shows performance at increasing pass@k thresholds as well as the overall success rate. These results highlight the nuanced effects of prompt style, reward granularity, and exploration on RL-based code generation in Q.
}

\label{fig:rl_ablation}
\end{figure}
\vspace{1em}

\vspace{1em}
\begin{figure}[h]
\centering
\includegraphics[width=\textwidth]{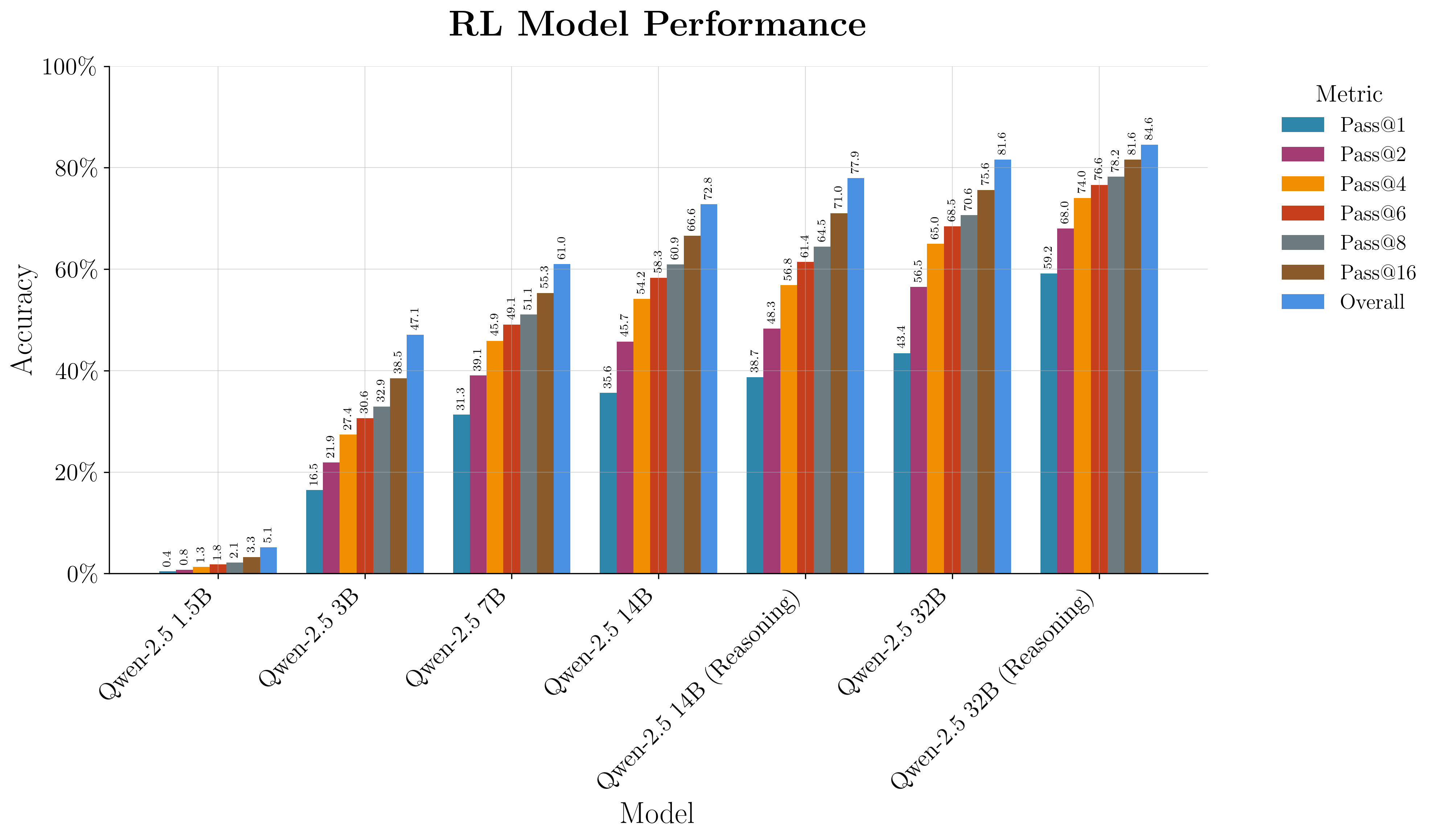}
\caption{
\textbf{RL Model Performance across Qwen-2.5 Series.}  
Pass@k results and overall accuracy for each Qwen-2.5 model after full adaptation, including reinforcement learning (RL). Bars indicate performance at different pass@k thresholds ($k=1,2,4,8,16$) as well as the overall percentage of problems solved. Both standard and reasoning-augmented models are shown for 14B and 32B. RL yields clear, consistent improvements with increasing model size, and the reasoning variants at the largest scales achieve the highest accuracies.
}
\label{fig:rl_full_results}
\end{figure}
\vspace{1em}

\begin{figure}[h]
    \centering
    \includegraphics[width=0.8\textwidth]{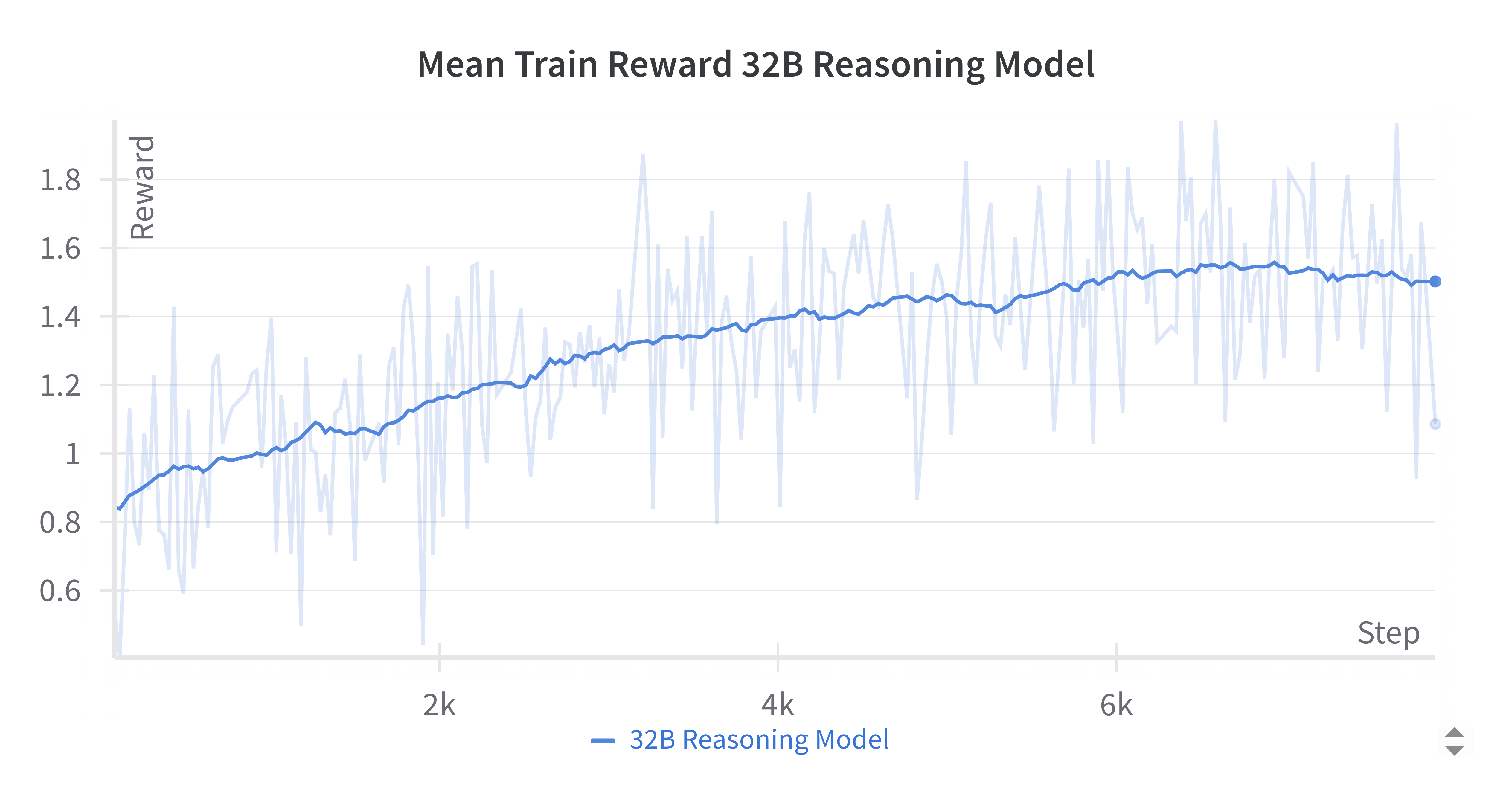}
    \caption{
    \textbf{Average training reward during GRPO for the Qwen-2.5-32B reasoning model.}
    The moving-average reward per training iteration is plotted across the full RL adaptation process. The curve exhibits a clear upward trend, indicating that the model steadily improved its ability to generate solutions that maximize the target reward, here, successful code completions on held-out evaluation problems. No significant instabilities or regressions were observed, demonstrating stable and effective reward-driven optimization at this scale.
    }
    \label{fig:32b_rl_reward}
\end{figure}

\begin{figure}[h]
    \centering
    \includegraphics[width=0.8\textwidth]{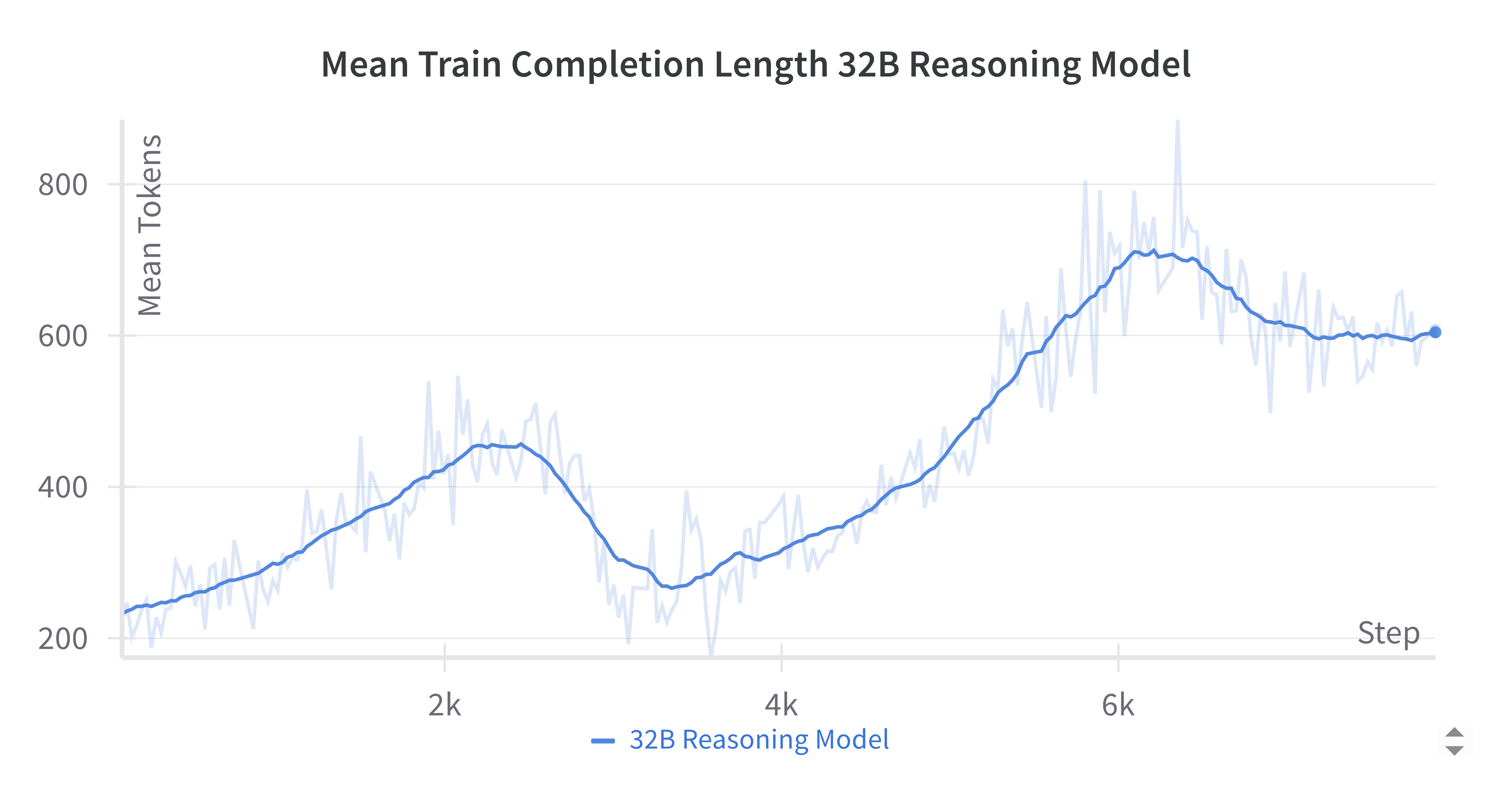}
    \caption{
    \textbf{Mean completion length during RL adaptation of the Qwen-2.5-32B reasoning model.}
    Average token count per generated solution as a function of RL training iteration. The observed non-monotonic trajectory, increase, decrease, then renewed increase, may indicate the model’s evolving approach to problem solving: from initial verbosity, through consolidation, to later-stage hybrid strategies. These shifts could reflect periods of exploration, exploitation, and renewed diversity as the policy adapts to the reward landscape.
    }
    \label{fig:32b_completion_length}
\end{figure}

\subsection{Setup and Training Protocol}
All RL experiments employed GRPO, a recent advancement in scalable RL for LLMs, implemented via Hugging Face’s TRL library \cite{vonwerra2022trl}. For practical scalability, we hosted all policy models on 7$\times$H100 GPUs using Accelerate \cite{accelerate} and DeepSpeed \cite{zero}, with a dedicated vLLM\cite{kwon2023efficient_vllm} inference server (on 1$\times$H100) to efficiently generate completions in parallel. This infrastructure enabled high-throughput, low-latency RL training and evaluation, critical for rapid experimentation.

For each RL run, we evaluated the policy on 15 LeetCode-Q problems per training round from our dataset, sampling four completions per problem, and updated policies every 25 steps. All RL experiments used a fixed learning rate of $2\times10^{-6}$, and were initialized from the best SFT checkpoint for each model size. All models were trained with the Adam optimizer.

Policy updates and sampling were tightly coupled: each RL process made API calls to the vLLM inference server for batch completion generation, allowing for rapid iteration and evaluation. To maximize GPU utilization and efficiency, training and inference were allocated to separate GPU sets, for example, with 8 GPUs, vLLM inference might run on GPU 0 while RL training ran on GPUs 1-7. The vLLM server shards model weights and handles incoming generation requests in parallel via tensor and data parallelism, ensuring all available inference GPUs remain fully utilized rather than idle. Meanwhile, the TRL trainer manages the RL process, including periodic synchronization of model weights: after each policy update, the updated weights are communicated back to the vLLM server to ensure inference reflects the latest policy, seamlessly integrating training and inference across devices. This orchestration allows for both fast online generation and efficient hardware usage without device contention.

\subsection{Ablation Dimensions and Reward Structure}
We systematically ablated several key axes of RL design, focusing on the 14B and 32B models (smaller models struggled to make meaningful progress under GRPO):

\begin{itemize}
\item \textbf{Reasoning vs non-reasonign models:} We compared \emph{reasoning} models (where the model is instructed to output intermediate reasoning or explanations before code) and \emph{non-reasoning} models (direct code output only). Both variants were trained and evaluated to assess the effect on solution accuracy and policy stability.
\item \textbf{Sampling Temperature:} We performed RL training with generation temperatures of 0.6, 0.7, 0.8, and 1.0. This allowed us to probe the exploration–exploitation tradeoff and discover whether diversity in sampling aids solution discovery or harms convergence.
\item \textbf{Reward Schemes:} We explored multiple reward signals, including:
\begin{itemize}
\item \emph{Test Case Reward:} The fraction of test cases passed by the generated Q code (0 for none, 1 for all, linear in between).
\item \emph{Perfect Bonus:} An additional bonus for solving all test cases (+2).
\item \emph{Combined Reward:} Mixtures of the above to probe sensitivity.
\end{itemize}
\end{itemize}

The use of programmatic feedback for reward, tied directly to our evaluation harness, enabled precise and scalable training without manual annotation or external judging. While controlling the length of reasoning has been shown to be important \cite{aggarwal2025l1controllinglongreasoning}, this is something we did not penalize for and leave for future work. 


\subsection{Results and Analysis}
RL with GRPO yielded measurable improvements for all model sizes except 1.5B, with the 14B and 32B models showing the most improvement. Particularly when using richer reward signals and moderate sampling temperatures. The reasoning trained model shows more improvement that normal prompting. Temperature ablations suggested that slightly higher diversity (up to $T=0.8$) could improve early-stage exploration, but excessively high temperatures harmed overall stability.

A complete set of ablation results is shown in Figure~\ref{fig:rl_ablation}, which breaks down the final reward and pass@k outcomes by experimental condition. The full experimental results can be seen in Figure~\ref{fig:rl_full_results}.

\subsection{Model Size and Prompting Effects}
RL adaptation on the smallest model (1.5B) did not yield meaningful improvements, in fact, under our GRPO setup, pass rates typically declined after RL, suggesting that limited capacity models may struggle to benefit from policy optimization in this setting. For the 14B model, we observed a nuanced relationship between prompting style and outcomes: reasoning-augmented models occasionally solved problems that non-reasoning variants failed to address, yet across the benchmark, non-reasoning models consistently achieved slightly higher overall pass@k rates. This suggests that explicit reasoning can enable novel solution strategies for difficult cases, but may also introduce additional complexity or variance that does not always translate into higher aggregate success. Whereas for the 32B model the reasoning model performed better across all pass@k rates and solved more problems overall.

\subsection{RL Training Dynamics for the 32B Reasoning Model}
To better understand the effects of reinforcement learning in our most capable setting, we analyze two key metrics from GRPO training of the 32B reasoning-augmented model: the average training reward per iteration and the mean completion length of generated solutions.

Figure~\ref{fig:32b_rl_reward} shows the moving-average training reward throughout RL. The overall trend is a steady and consistent improvement, reflecting the model’s increasing alignment with the reward signal and its growing ability to generate successful solutions under the specified evaluation criteria.

In Figure~\ref{fig:32b_completion_length}, we plot the mean length (in tokens) of model-generated completions over the course of RL training. Interestingly, this metric displays a non-monotonic trajectory: average completion length increases initially, then decreases, before trending upward again in later iterations. This pattern may reflect the model’s exploration and refinement of solution strategies, starting with more verbose reasoning, later converging on concise forms, and finally incorporating a mix of both as new policies are explored. The dynamic adaptation of completion length highlights the model’s flexibility in approaching the problem and suggests that different stages of training emphasize distinct reasoning or coding styles.

\section{Final Results}
\begin{figure}[t]
\centering
\includegraphics[width=\textwidth]{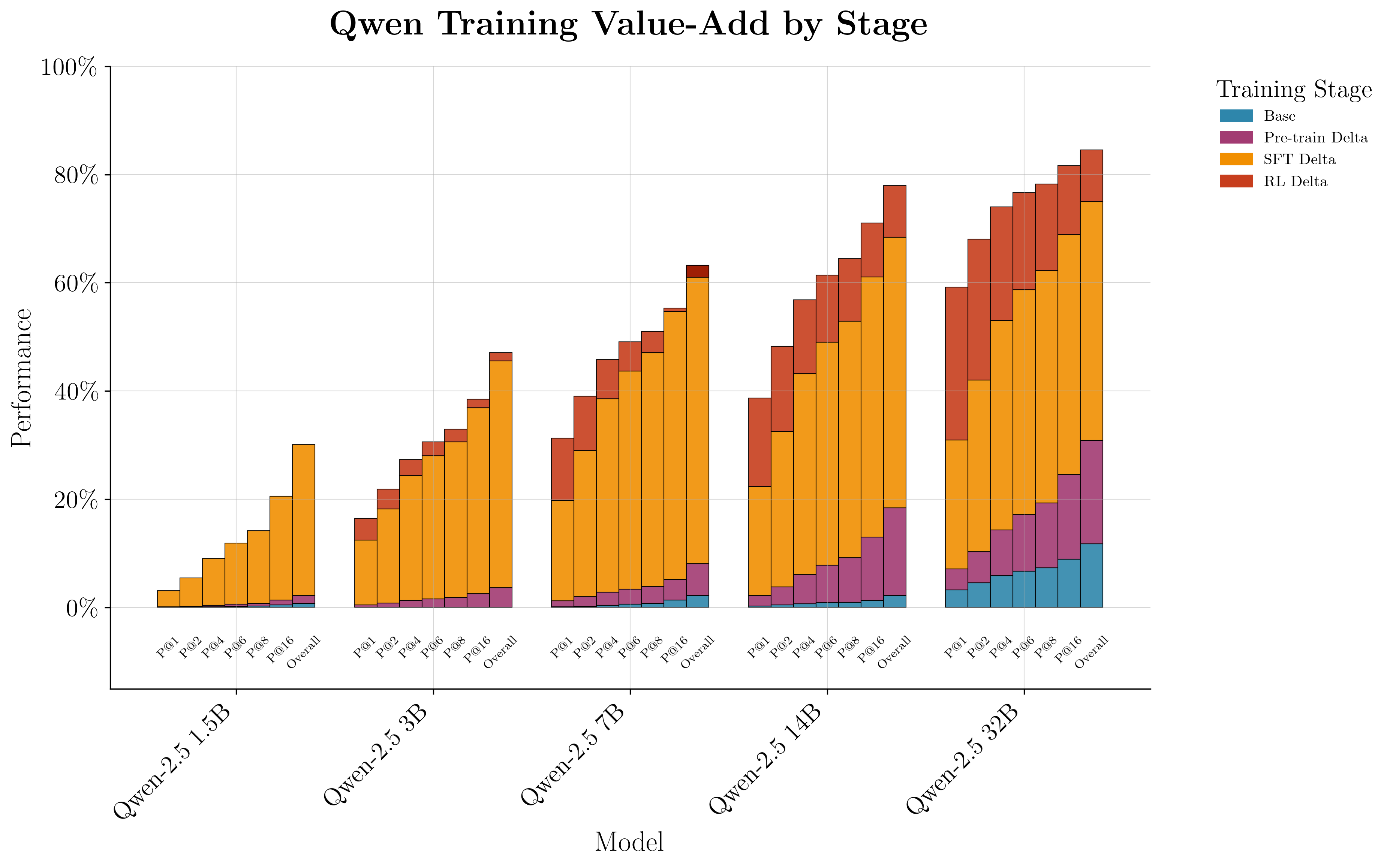}
\caption{
\textbf{Cumulative gains across adaptation stages for each model.}
Stacked bar chart showing pass@k improvements for all Qwen-2.5 model sizes, at each step: base (benchmark), after pretraining, after SFT, and after RL. For 1.5B, RL did not yield further gains and is omitted; for 14B and 32B, we report the best reasoning-model results. These results highlight how each adaptation stage builds on the previous, with significant gains achieved for larger models.
}
\label{fig:final_stacked_results}
\end{figure}

Having walked through the full pipeline, from benchmarking foundation models to domain-adaptive pretraining, supervised fine-tuning, and reinforcement learning, we can now summarize the end-to-end improvements. Figure~\ref{fig:final_stacked_results} visualizes the cumulative gains across each stage for all Qwen-2.5 model sizes, highlighting the contribution of each method.

\vspace{1em}

To directly compare all models, Table~\ref{tab:delta_to_best} shows the difference in pass@k rates between each Qwen-2.5 variant and the best overall model (Opus-4) on the Q-LeetCode test set.

\begin{table}[h]
\centering

\begin{tabular}{lcccccc}
\toprule
\textbf{Model} & \textbf{pass@1} & \textbf{pass@2} & \textbf{pass@4} & \textbf{pass@8} & \textbf{pass@16} & \textbf{Overall} \\
\midrule
\textbf{Qwen-2.5 1.5B} & -29.2\% & -36.1\% & -42.3\% & -47.9\% & -53.2\% & -58.8\% \\
\textbf{Qwen-2.5 3B} & -13.2\% & -15.0\% & -16.2\% & -17.2\% & -18.0\% & -16.9\% \\
\textbf{Qwen-2.5 7B} & +1.7\% & +2.1\% & +2.3\% & +1.0\% & -1.2\% & -2.9\% \\
\textbf{Qwen-2.5 14B} & +9.0\% & +11.4\% & +13.2\% & +14.4\% & +14.6\% & +14.0\% \\
\textbf{Qwen-2.5 32B} & +29.5\% & +31.1\% & +30.4\% & +28.1\% & +25.1\% & +20.6\% \\
\bottomrule
\end{tabular}
\caption{Difference in pass@k rates ($k=1,2,4,8,16$) and Overall Success Rate between each Qwen-2.5 model (after full adaptation) and the best-performing model (Opus-4) on the Q-LeetCode benchmark. Negative values indicate a gap to the best model.}
\label{tab:delta_to_best}
\end{table}

\vspace{1em}

Taken together, these results demonstrate that our staged adaptation process enabled substantial improvements over baseline models, with the largest Qwen-2.5 models approaching, but not yet matching, the performance of leading closed-source LLMs. Our pipeline offers a clear blueprint for effective domain adaptation in low-resource and specialized domains.

\section{Hardware and Tools}
All experiments in this work were made possible by the outstanding open-source ecosystem surrounding large language models. Our training and evaluation pipelines were built primarily on the Hugging Face \texttt{transformers} \cite{wolf-etal-2020-transformers} and \texttt{trl} \cite{vonwerra2022trl} libraries, which enabled rapid iteration and reproducibility for both supervised and reinforcement learning workflows. Experiment tracking and artifact management were handled entirely through Weights \& Biases (W\&B) \cite{wandb}, providing transparent logs, metrics, and model checkpoints for all runs.



Computation was performed on a dedicated cluster of sixteen NVIDIA H100 GPUs on  \href{https://www.coreweave.com/}{CoreWeave}. CoreWeave’s AI Cloud Platform simplified the complexity of running and monitoring state-of-the-art AI infrastructure and delivered high scale and reliability to power our research. This high-performance environment inclusive of compute, network-ing, storage, and observability allowed us to scale experiments across multiple Qwen-2.5-Instruct model sizes, from 1.5B to 32B parameters, and to explore a wide range of ablation and reinforcement learning settings without resource bottlenecks. All multi-GPU training leveraged DeepSpeed ZeRO \cite{zero} Stage 3 for efficient memory partitioning and scaling, with distributed orchestration via Hugging Face Accelerate. Inference serving for RL and batch evaluation was handled by the vLLM \cite{kwon2023efficient_vllm} engine, enabling high-throughput, low-latency sampling during training loops. 

All code, configuration files, and environment details are included in our public re-
lease to facilitate reproduction and extension. We gratefully acknowledge the support
of CoreWeave and the open-source community in making this research possible.

\section{Limitations}
While our work demonstrates a viable pipeline for domain adaptation and benchmarking with Q code, it is important to recognize the limitations of our current approach. The LeetCode-derived dataset used in this study is intentionally “pythonic” in nature: it frames Q as a target for algorithmic problem-solving and code translation tasks more common in Python or other general-purpose languages, rather than reflecting the typical usage patterns of Q in real-world settings. In practice, Q is most often used for high-performance querying, analytics, and time-series operations, domains that are fundamentally different from LeetCode-style algorithmic challenges.

To clarify the distinction between typical real-world Q usage and the LeetCode-style Q code used in our benchmark, Figure~\ref{fig:q_style_comparison} presents representative examples of each. As shown, real-world Q emphasizes expressive, high-performance queries and analytics, while our LeetCode-derived dataset frames Q as an algorithmic problem-solving language in a Pythonic style. This side-by-side comparison illustrates the gap between practical Q deployment and the current evaluation benchmark.

\begin{figure}[h]
\centering

\begin{minipage}[t][7.8cm][t]{0.47\textwidth}
\begin{tcolorbox}[colback=black!1!white, title=Real-World Q (Query/Analytics), boxsep=4pt, left=2pt, right=2pt, height=7cm, valign=top, top=0pt]
\begin{lstlisting}[language={},basicstyle=\ttfamily\footnotesize]
// Select trades for IBM on the most recent date
t: select time, price from trade
    where date = last date, sym = `IBM

// Calculate average trade size per symbol
avgSz: select sym, avg size by sym from trade
\end{lstlisting}
\end{tcolorbox}
\end{minipage}
\hspace{0.03\textwidth}
\begin{minipage}[t][7.8cm][t]{0.47\textwidth}
\begin{tcolorbox}[colback=black!1!white, title=LeetCode-Style Q (Algorithmic), boxsep=4pt, left=2pt, right=2pt, height=7cm, valign=top, top=0pt]
\begin{lstlisting}[language={},basicstyle=\ttfamily\footnotesize]
// Check if a list is a palindrome
isPal:{[x] x ~ reverse x}

// Calculate the H-Index from a list of citations
solve:{[citations]
  n:count citations;
  citations:asc citations;
  i:n;
  while[i>0;
    if[citations[n-i] >= i; :i];
    i:i-1];
  :0}
\end{lstlisting}
\end{tcolorbox}
\end{minipage}

\caption{Comparison between typical real-world Q code (left), focused on querying and analytics, and LeetCode-style Q code (right), which frames Q as an algorithmic problem-solving language.}
\label{fig:q_style_comparison}
\end{figure}

As such, we do not claim that our models are strong Q practitioners in the true sense, nor do we suggest that high pass rates on LeetCode-Q directly translate to real-world impact for Q developers. Instead, we view this work as a blueprint and base model: a resource for others to adapt and build upon, whether for Q or for other niche domains where evaluation and training data are scarce. Our models may serve as useful starting points for further domain-specific fine-tuning, especially as better datasets and benchmarks for real-world Q usage become available.

Developing benchmarks and models that more faithfully reflect the SQL-like, database-focused style of practical Q code is a priority for future research. We hope this release will serve both as a foundation and as encouragement for the community to push further in this direction.

\section{Artifacts}
All models developed in this work are released publicly on Hugging Face. For general-purpose Q usage, we recommend starting with either the 32B reasoning model or the fully pretrained checkpoint, as these demonstrated the best overall performance in our experiments. However, users can experiment with and benchmark all available model variants, from 1.5B up through 32B, to suit their resource constraints and application needs.

In addition to the models, we release the complete codebase used for dataset preparation, training, evaluation, and reinforcement learning on GitHub. All scripts, configuration files, and detailed instructions are provided to enable full reproduction of our results and to facilitate further research and development in Q or other niche domains.

\section{Lessons Learned}
While many of the lessons from this project echo common wisdom in machine learning and LLM research, our experience brought them into sharp focus:

\begin{enumerate}
\item \textbf{Everything hinges on evaluation.}
Your evaluation harness is the foundation of the entire project, and it’s where you have the most direct control. Invest in conceptually clear metrics, even if it means narrowing your scope to a single, well-defined axis of improvement. Strive for fully programmatic and deterministic evaluation wherever possible. If you must use an LLM as a judge (even for simple verification), rigorously validate its reliability, use multiple rounds or explicit rubrics, and ensure you’re genuinely measuring what you care about. From an engineering perspective, abstract all model calls through an API, and build for high parallelism. Hosting your own models with vLLM \cite{kwon2023efficient_vllm} (or similar engines) from the start adds tremendous agility, enabling rapid and scalable benchmarking and RL.

\item \textbf{Fast inference matters.}  
Use vLLM \cite{kwon2023efficient_vllm} or equivalent high-throughput inference engines for evaluation and RL. This dramatically accelerates experimentation and unlocks smoother integration of both frontier and self-hosted models.

\item \textbf{Data quality pays off.}  
Time spent curating, cleaning, and filtering your dataset is always repaid in model quality and reliability. Even modest improvements in data fidelity often translate into outsized gains downstream.

\item \textbf{Reward hacking is both pervasive and subtle.}  
Reward hacking is not just a theoretical concern for reinforcement learning; it can manifest in any stage where model outputs are selected or filtered based on imperfect evaluation criteria. We repeatedly observed models exploiting loopholes in our evaluation harness, most strikingly when solution and test case generation were insufficiently separated, allowing models to “game” the process by producing trivial or overfitted test cases. While this is well-known in RL (where models will invariably find shortcuts to maximize reward), it is equally relevant in rejection sampling and supervised fine-tuning (SFT) pipelines, where accepted samples reflect the biases or gaps in your filtering logic. In tasks with programmatic, verifiable metrics, reward hacking can be minimized by ensuring strict independence between solution and evaluation, and by using robust multi-case test suites. However, in softer-reward or less rigorously specified tasks, it becomes much more challenging, careful validation, adversarial testing, and periodic manual review remain essential safeguards.

\item \textbf{Large models are essential for meaningful learning.}  
Through extensive experimentation, we found that genuine progress in Q domain adaptation only emerged with models of at least 14B parameters, and the 32B model in particular demonstrated qualitatively superior capabilities. At this scale, the model exhibits robust world knowledge, strong instruction following, and shows clear benefits from reinforcement learning, behaving much more like a “real” assistant than smaller variants. While smaller models (e.g., 1.5B or 3B) may suffice for highly constrained or well-defined tasks, the gap in generalization and reasoning is substantial. For ambitious goals, a 32B model represents a practical threshold for impressive performance.

\item \textbf{Training large models is accessible on single-node hardware, but scaling out accelerates research.}
Although 32B parameters may sound prohibitive, we found that all phases, pretraining, supervised fine-tuning, and RL adaptation, can be run on a single 8$\times$H100 GPU node using efficient frameworks like Accelerate \cite{accelerate} and DeepSpeed \cite{zero}. For example, a single supervised fine-tuning (SFT) run on the 32B model typically required about 12 hours on 8 H100 GPUs. However, scaling out to multi-node clusters provided dramatic speedups for large-scale experimentation. In practice, we utilized up to 16 H100 GPUs in parallel for hyperparameter sweeps and ablations, allowing us to halve our experimental time. On occasion, we pulsed up to 64 GPUs via SLURM for the largest jobs, which further accelerated research turnaround for ambitious experiments.

\item \textbf{Meticulous experiment tracking is a must.}  
With the exponential number possible configurations, hyperparameters, and evaluation protocols, the value of robust experiment management cannot be overstated. Using tools like Weights \& Biases \cite{wandb} from the start, and agreeing on consistent logging and formatting conventions, proved essential for making sense of results, facilitating collaboration, and ensuring reproducibility. A clean, searchable experiment log is as important as the code or models themselves.

\item \textbf{Training is the easy part, if your eval and data are strong.}  
With well-designed datasets and robust evaluation, model training, especially for pretraining and SFT, is often straightforward. Learning rate sweeps and early stopping were sufficient for consistently good results. RL was more nuanced, but a strong evaluation pipeline made troubleshooting and iteration much more manageable.

\item \textbf{Prove your pipeline on the simplest task first.}  
Start with the minimal working example: select the clearest, most verifiable axis for early progress. Once the pipeline is validated and reliable end-to-end, extension to more complex settings is much easier and less error-prone.

\item \textbf{Start simple with reinforcement learning.}  
Reinforcement learning (RL) remains the trickiest and most sensitive part of LLM training pipelines. It can be difficult to know which hyperparameters matter most, or even whether the overall setup is functioning as intended. Our strong recommendation is to begin with the simplest possible reward signal, ideally one that is unambiguous and easy to optimize, such as maximizing the number of vowels in a completion. This “sanity check” ensures that the RL loop, optimizer, and reward computation are wired up correctly, and that your model is capable of learning at all. In our experience, failing to do this can lead to wasted cycles diagnosing “learning problems” that are actually pipeline or configuration errors (e.g., overly high learning rates, misapplied gradients, or subtle bugs). Only once a toy reward works should you proceed to the full, domain-specific setup.

\end{enumerate}

Overall, these principles kept the project on track, guided our engineering and research decisions, and provided a repeatable path to tangible improvement.

\section{Conclusion}
Large language models have shown remarkable capabilities, but their true value emerges when adapted and specialized for the domains that matter most. While this process can seem daunting, especially for niche languages like Q, we have aimed to provide a transparent blueprint: from building a usable dataset and evaluation harness, to benchmarking foundation models, to domain-adaptive pretraining, supervised fine-tuning, and reinforcement learning.

We hope this work serves as a practical case study for the broader community, illustrating how to approach and organize the fine-tuning process in low-resource settings. Our open-source code and artifacts are designed to lower the barrier for others to experiment and build. For the Q programming community, our released models, especially the pretrained and reasoning-optimized variants, offer a new starting point for both general and specialized applications, or as foundations for further adaptation.

Ultimately, we believe that the combination of robust evaluation, iterative improvement, and open research will continue to drive progress, not just for Q, but for all specialized domains where LLMs can unlock new value.

\section{Acknowledgments}
We would like to thank Aaron Davies, Sebastian Dragnea, Ross Duffy, Alvin Shih, Stephen Tse, Kalyan Vadlakonda, Matthew Clark, and Johannes Palmgren for testing the model. We are also grateful to Oana Frunza, Sanchit Sinha, Alfonso Amayuelas Fernandez, Adit Jain, and Chuanyang Zheng for their thoughtful reviews and feedback on our draft.

We are especially thankful to CoreWeave, whose purpose-built AI cloud platform  powered all of our experiments. In particular, we would like to thank Ethan Walton and Alan Zaccone from CoreWeave for their generous support throughout this project.

We would like to also thank Larry Bromberg, Max Iori, and the Morgan Stanley Innovation Lab team for their support of our work. 

Finally, we extend our thanks to Thomas Kamei and John Haggarty for their insightful feedback.

\newpage
\bibliographystyle{plain}
\bibliography{references}

\newpage
\appendix
\appendixpage    
\addappheadtotoc

\section{Reasoning Models}
To assess the impact of explicit reasoning and step-by-step thinking on Q code generation, we evaluated leading “reasoning” models, including Grok-4, Gemini 2.5 Pro and OpenAI’s o3, under a constrained budget. Each model was given a maximum of 2,000 total output tokens across all test problems, reflecting both practical runtime concerns and our belief that LeetCode problems, especially in a non-mainstream language like Q, are better served by models that are simply aware of the syntax rather than relying on extended reasoning chains. For Grok-4, this token budget yielded 630 completions; for Gemini 2.5 Pro 53 completions and for o3, 5,438 completions.

\begin{figure}[h]
\centering
\includegraphics[width=\textwidth]{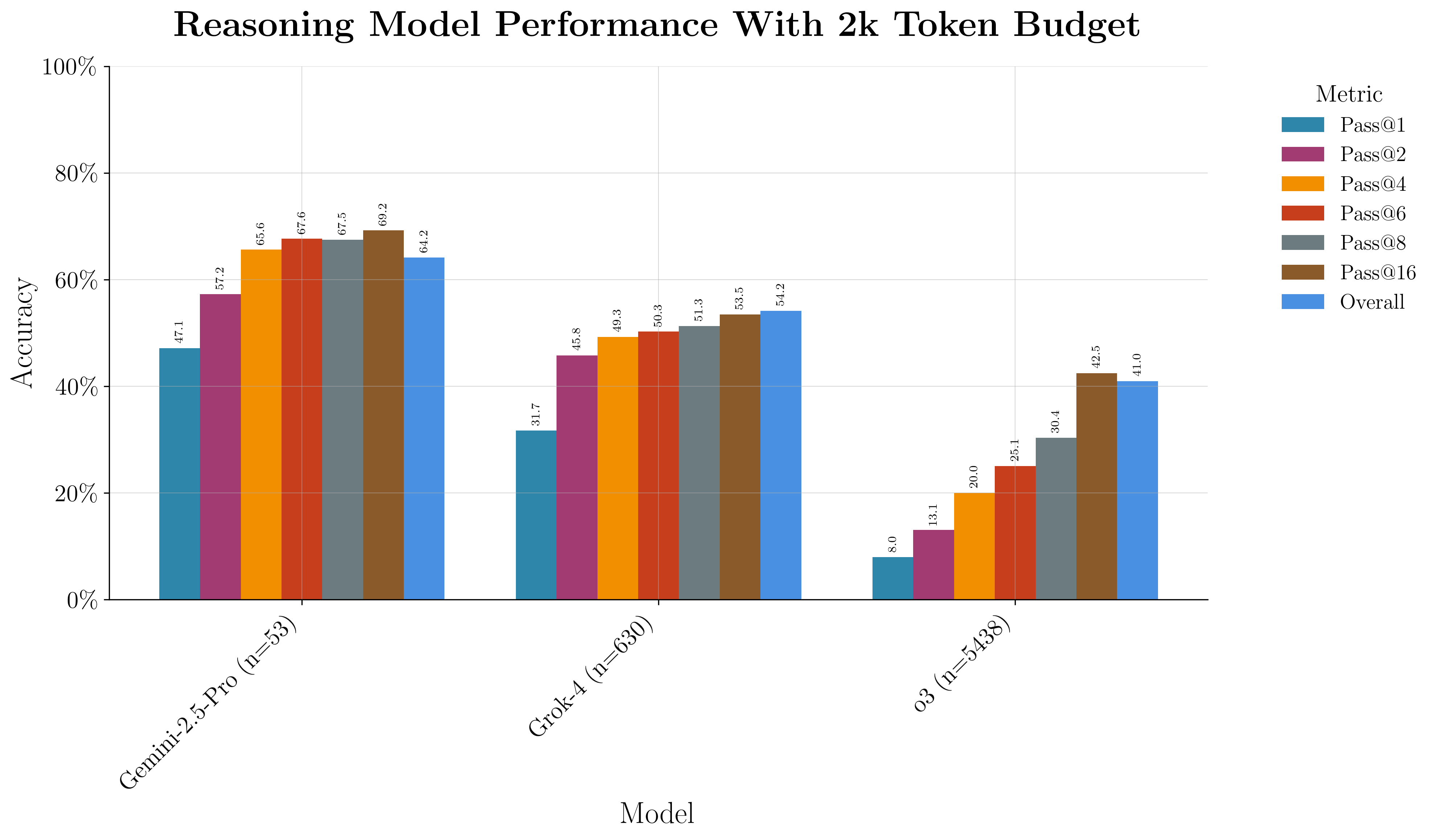}
\caption{
\textbf{Performance of reasoning-oriented models under constrained token budgets.}
Results for Grok-4, Gemini-2.5 Pro and o3 are shown, where each model was allowed a total of 2,000 generated tokens for the entire Q-LeetCode benchmark. Performance is reported only on the subset of completions produced within this budget (53 for Gemini-2.5-Pro, 630 for Grok-4, 5,438 for o3). We chose not to run unconstrained “reasoning” settings, as our dataset primarily evaluates Q syntax awareness on standard coding tasks, contexts that do not inherently require lengthy step-by-step reasoning.
}
\label{fig:reasoning_models}
\end{figure}

\section{Category-Level Error Analysis}
To better understand the strengths and weaknesses of both the API models and the Qwen-2.5 base models, we analyzed model errors by LeetCode problem category and problem hardness. These breakdowns highlight where current models tend to struggle and can inform targeted future improvements.

\begin{figure}[h]
\centering
\includegraphics[width=\textwidth]{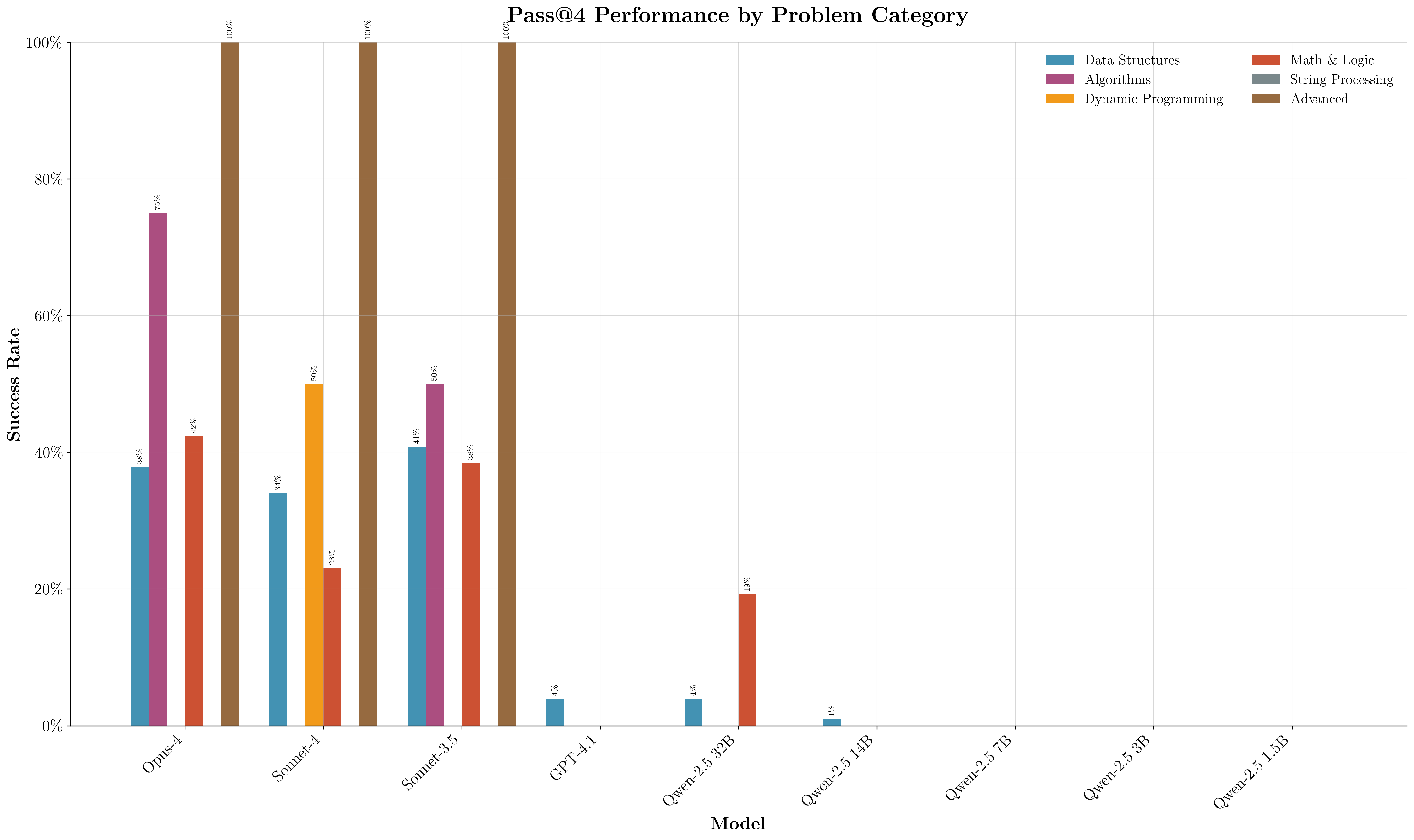}
\caption{
\textbf{Per-category error distribution for API models and Qwen base models.}
Each bar shows the number of problems missed within each LeetCode problem category (e.g., Array, Hash Table, Dynamic Programming). This breakdown illustrates the distribution of model errors and suggests which problem types may require more specialized attention for future adaptation.
}
\label{fig:category_error_analysis}
\end{figure}

\begin{figure}[h]
\centering
\includegraphics[width=\textwidth]{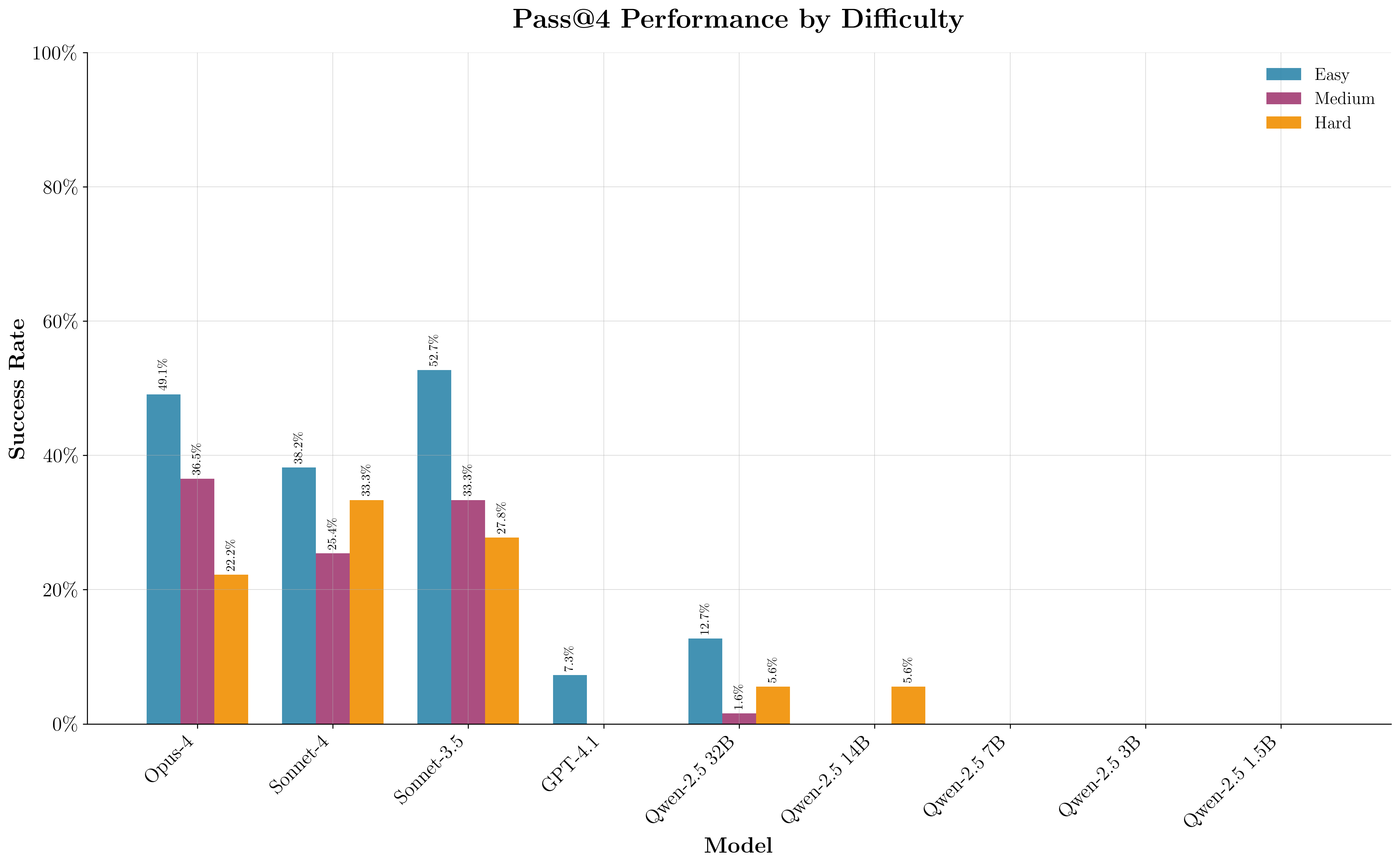}
\caption{
\textbf{Error breakdown by problem difficulty.}
Model performance stratified by LeetCode difficulty (Easy, Medium, Hard) for both API and Qwen base models. This plot demonstrates how model accuracy declines with increasing problem hardness, and provides insight into where further improvements are most needed.
}
\label{fig:hardness_error_analysis}
\end{figure}

\section{Prompts Used in Data Generation and Training}
For reproducibility and future research, we provide the exact prompts used for data generation, supervised fine-tuning, and reinforcement learning, including both reasoning-augmented and non-reasoning variants. All prompts are formatted as used in the actual pipeline. Theses can also all be found in the Github. 

\vspace{2em}

\subsection{Data Generation Prompt}

\begin{tcolorbox}[breakable, colback=white!98!gray, colframe=black!60!white,
    title=\textbf{Prompt: Q Code Generation from LeetCode Problem}]
\begin{lstlisting}[basicstyle=\ttfamily\small, breaklines=true]
# Translating Solutions: 
Translate the following Python code to Q programming language.

PROBLEM CONTEXT:
{problem_desc}

PYTHON CODE TO TRANSLATE:
{python_code}


Please provide only the Q code (no markdown, no explanations):

# Translating Test Cases: 
Convert this Python test case to Q format:

Python test: print(solve({args}))
Expected output: {expected_output}

Example Q test format:
result:solve[58];
show result;

result:solve enlist ("dog "; "racecar "; "car ");
show result;

result:solve[(0,1,1)];
show result;

Q SYNTAX NOTES:
- Call function: result:solve[args];
- Display result: show result;
- Lists in Q: (item1;item2;item3)
- Strings: "text"
- For single-item lists use 'enlist': enlist "item"

CRITICAL REQUIREMENT: Your Q test case MUST include a call to the solve function. The first line must be something like "result:solve[...];" where you call the solve function with the appropriate arguments.

Provide only the Q test code (2 lines: assignment and show):
\end{lstlisting}
\end{tcolorbox}
\vspace{1em}

\begin{tcolorbox}[breakable, colback=white!98!gray, colframe=black!60!white,
    title=\textbf{Supervised Fine-Tuning (SFT) Prompt}]
\begin{lstlisting}[basicstyle=\ttfamily\small, breaklines=true]
# Description to Q 
Write a Q solve function that solves the following problem:\n\n{desc}\n\nOutput ONLY the Q solve function. Do not include any other text, explanations, or a test harness.

# Python to Q
"Translate the following Python `solve()` function to an equivalent Q `solve` function.\n\nPython `solve()` function:\n```python\n{py_code}\n```\n\nOutput ONLY the Q `solve` function. Do not include any other text, explanations, or a test harness.

# Q to Python
Translate the following Q `solve` function to an equivalent Python `solve()` function.\n\nQ `solve` function:\n```q\n{q_code}\n```\n\nOutput ONLY the Python `solve()` function. Do not include any other text, explanations, or a test harness.

\end{lstlisting}
\end{tcolorbox}
\vspace{1em}

\begin{tcolorbox}[breakable, colback=white!98!gray, colframe=black!60!white,
    title=\textbf{Reinforcement Learning (RL) Prompts}]
\begin{lstlisting}[basicstyle=\ttfamily\small, breaklines=true]
# Translating Solutions: 
Your task is to write correct Q code that solves the given programming problem.
Problem: 
{problem_description}

Use this exact format for your response:
<reasoning>
Explain your approach here
</reasoning>

<answer>
// Your Q code here
</answer>


\end{lstlisting}
\end{tcolorbox}
\vspace{1em}

\vspace{1em}

\end{document}